\documentclass[11pt]{article}

\usepackage[utf8]{inputenc}
\usepackage[T1]{fontenc}
\usepackage[english]{babel}
\usepackage{lmodern}
\usepackage{microtype}
\usepackage[margin=1in]{geometry}
\usepackage{graphicx}
\usepackage{booktabs}
\usepackage{siunitx}
\usepackage{amsmath}
\usepackage{multirow}
\usepackage{enumitem}
\usepackage{float}
\usepackage[section]{placeins}   
\usepackage{xcolor}

\usepackage[numbers,sort&compress]{natbib}
\usepackage[colorlinks=true,linkcolor=black,citecolor=black,urlcolor=blue]{hyperref}

\graphicspath{{figures/}}

\title{\textbf{Do Large Language Models Know Colombian Law?\\
A Reliability Benchmark for the Colombian Legal System}}

\author{
  Rub\'en Manrique$^{1}$ \and
  Michelle Castellanos$^{2}$ \and
  Jorge Morales$^{1}$ \and
  Juan David Guti\'errez$^{2}$ \and
  Antonio Barreto Rozo$^{3}$ \and
  Joaqu\'in V\'elez Navarro$^{3}$
}
\date{%
  \small $^{1}$Department of Systems and Computing Engineering,
  Universidad de los Andes\\
  \small $^{2}$School of Government, Universidad de los Andes \quad
  $^{3}$Faculty of Law, Universidad de los Andes\\[4pt]
  \today
}

\begin{document}
\maketitle

\begin{abstract}
\noindent Large language models (LLMs) are increasingly used to support legal
practice, education, and research, yet their reliability in national legal
systems outside the United States remains largely undocumented. We introduce an
expert-validated benchmark for evaluating LLM reliability on the Colombian legal
system. The benchmark comprises \num{1042} items spanning ten areas of law and
three question formats---closed single-best-answer multiple-choice, semi-open,
and open-ended (IRAC)---built through a human-in-the-loop pipeline with
multi-stage expert review. We evaluate
\num{15} contemporary proprietary and open-weight models, scoring each question
format with format-appropriate metrics. Accuracy on closed questions ranges
widely, from \num{0.905} (Gemini~3.1~Pro) to \num{0.577}; but on free-text legal
answers, factual correctness never exceeds \num{0.45} for any model. Both metrics
range from \num{0} to \num{1}, where \num{1} denotes a perfect score. We find a
dissociation between answer relevancy and correctness (Spearman
$\rho=-0.46$), so that models reliably \emph{sound} responsive while frequently
being \emph{wrong}, a pattern of particular concern for non-expert users. Closed-question
accuracy and free-text correctness are strongly rank-correlated
($\rho=0.94$), so cheap multiple-choice screening predicts model ranking but
overstates absolute reliability. The free-text ranking proves robust, because an
independent rubric-based LLM judge and blind human expert scoring both reproduce
it ($\rho\geq0.88$). That judge, validated against the human experts, further reveals
that the norms models cite are frequently unreliable---only about half are
correct, and the rest cite the wrong norm or one that does not exist. Reliability
varies systematically by legal area and follows an inverted-U across question
complexity. Our results indicate that
current LLMs require expert supervision for Colombian legal tasks, and that
grounding answers in authoritative sources is a promising path to higher
reliability. We release the benchmark construction pipeline to support
reproducible evaluation.
\end{abstract}

\section{Introduction}
\label{sec:intro}
Large language models (LLMs) have moved quickly from novelty to a working tool in
the legal profession. A 2023 UNESCO global survey found that \num{40}\% of
surveyed judges, lawyers, and other judicial operators had already used ChatGPT
or similar tools for work, most often to search for legal information
(case law, statutes, doctrine) and to draft documents \citep{unesco2024judges}.
No comparable survey exists for Colombia, but the Consejo Superior de la
Judicatura\footnote{The Consejo Superior de la Judicatura is the body in charge
of administering the Colombian judicial branch.} found that roughly one in three
of the \num{3152} judicial employees
it surveyed had used AI systems, ChatGPT being the most common
\citep{csdj2025}. Adoption, in short, is already widespread and largely
unsupervised.

These tools, however, exhibit a well-documented failure mode. LLMs
\emph{hallucinate}, producing fluent but incorrect or fabricated legal content
\citep{dahl2024}. Lawyers and judges in several countries have pasted
tool-generated text---including citations to non-existent statutes and
case law---into briefs and judicial decisions without verification
\citep{gutierrez2024chapter,unesco2025guidelines}. Empirical studies in the
United States quantify the risk. General-purpose models such as GPT-4
hallucinate on more than half of legal queries and struggle to recognize their
own errors \citep{dahl2024}, and even commercial tools marketed as
``hallucination-free,'' such as Lexis+~AI and Westlaw AI-Assisted Research,
answer incorrectly between \num{17}\% and \num{33}\% of the time despite using
retrieval-augmented generation (RAG) \citep{magesh2025}.

Almost all of this evidence concerns English-language, U.S. common law. Empirical
research on LLM reliability in Spanish-speaking, civil-law jurisdictions---and
on Colombian law in particular---remains scarce; a recent systematic review finds
no dedicated Spanish-language legal QA benchmark for Latin America
\citep{manrique2026mindgap}. The gap matters because
linguistic, terminological, and normative differences mean that findings obtained
in English do not transfer automatically to Spanish legal practice, and commercial AI
systems are not typically representative of Latin American legal diversity. The
one prior assessment for Colombia concluded that, as of 2024, no evaluated model
reached an accuracy acceptable for practical legal use \citep{arevalo2024}, but
it predates the current generation of frontier models and covered a very limited
set of areas and question types.

We address this gap with a systematic, expert-validated evaluation of LLM
reliability on the Colombian legal system. Our contributions are:
\begin{itemize}
  \item \textbf{An expert-validated Colombian legal benchmark} of \num{1042}
  items spanning ten areas of law and three question formats---closed
  single-best-answer multiple-choice, semi-open, and open-ended (IRAC)---built through a
  human-in-the-loop pipeline with multi-stage expert review
  (Sections~\ref{sec:dataset}--\ref{sec:methodology}).
  \item \textbf{A reproducible evaluation of \num{15} contemporary LLMs},
  proprietary and open-weight, scored with format-appropriate metrics computed
  over explicit question subsets (Section~\ref{sec:setup}).
  \item \textbf{An analysis of where and how models fail} (Section~\ref{sec:results}),
  including a dissociation between answer relevancy and correctness---models
  reliably \emph{sound} responsive while frequently being \emph{wrong}---and a
  characterization of reliability by legal area and question complexity.
\end{itemize}

\section{Related Work}
\label{sec:related}

\subsection{Legal hallucinations and the reliability of LLMs}
A growing body of literature measures how often LLMs produce incorrect legal content.
\citet{dahl2024} provide the first systematic profile of legal hallucinations in
public-facing models, reporting hallucination rates above \num{50}\% on a range
of tasks and a typology of failure modes. \citet{magesh2025} extend this to
commercial, RAG-based legal research tools, showing that ``hallucination-free''
claims are overstated and that error rates of \num{17}--\num{33}\% persist.
Complementarily, \citet{guha2023legalbench} introduce LegalBench, a
collaboratively built benchmark for legal reasoning, establishing the value of
task-structured, expert-informed evaluation. For open-ended legal answers, where
correctness is not a single string, \citet{salinas2026lawprofs} conduct a blinded
study in which sixteen U.S. law professors rate LLM and peer short answers; the
models are preferred over human peers, and the authors show that a separate LLM
acting as a judge reproduces expert agreement, supporting LLM-as-a-judge as a
scalable, expert-aligned evaluation method for judgment-rich legal tasks. Our
work adopts this evaluation philosophy---and a reference-anchored LLM-judge
protocol in the same spirit---but targets a non-U.S., civil-law jurisdiction and
explicitly separates recognition (closed questions) from generation (free text).

\subsection{AI in the Justice System in Latin America and Colombia}
A parallel literature documents the adoption and governance of AI in judicial
and public-sector settings. Global and regional surveys map where and how
judicial operators use these tools \citep{unesco2024judges,csdj2025,cepej2025},
and guidance has begun to emerge on their responsible use in courts
\citep{unesco2025guidelines,gutierrez2020}. In Colombia specifically, studies
have examined automated decision-making in the public sector
\citep{gutierrez2025adm,florez2020}, the use of AI in the Attorney General's
Office \citep{palacios2024fiscal}, and the Constitutional Court's PretorIA system
for processing human-rights cases \citep{saavedra2021pretoria}. Critical
appraisals caution against uncritical reliance on LLMs in judicial
decision-making \citep{gutierrez2024chapter,barreto_inpress}. This body of work
establishes the stakes and the demand for AI in Colombian legal practice; it
motivates, but does not itself provide, a systematic measurement of how reliably
current LLMs actually answer Colombian legal questions.

\subsection{Spanish-language legal QA in Latin America}
The work that most directly motivates this benchmark is the systematic review and
zero-shot evaluation of \citet{manrique2026mindgap}. Surveying \num{41} legal-NLP
benchmarks published between 2018 and 2026, they find that Latin America is
represented by only six---all covering Brazil and Portuguese---and that
\emph{no} dedicated Spanish-language legal QA benchmark exists, despite Spanish
being spoken across eighteen civil-law countries in the region. They further
provide a first zero-shot evaluation of five LLMs on \num{458} bar-exam-style
multiple-choice questions from Bolivia, Colombia, and Puerto Rico, where the best
model reaches only \num{79.7}\% accuracy and performance is weakest in criminal
and administrative law. Crucially for the present work, that study is confined to
\emph{closed} multiple-choice items and explicitly calls for dedicated
Spanish-language legal benchmarks that go \emph{beyond} the multiple-choice format
and incorporate expert validation. Our benchmark is a direct response to that
call: it focuses on Colombian law in depth, adds semi-open and open-ended (IRAC)
formats alongside closed questions, applies a multi-stage expert-validation
protocol, and evaluates a larger and more recent set of models.

An earlier and narrower precedent is \citet{arevalo2024}, who evaluate three LLMs
(ChatGPT-3.5, ChatGPT-4, and Gemini Pro) on Colombian legal queries across five
areas of law and find that ChatGPT-4 performed best overall but that no model
reached an acceptable level of accuracy for practical use, without releasing a
reusable benchmark. Relative to both, we evaluate a larger and more diverse set
of \num{15} contemporary models on an expert-validated benchmark that is larger
(\num{1042} items), broader (ten legal areas), and stratified by question format and
complexity, and we report metrics matched to each format---separating factual
correctness from surface relevancy---which reveals reliability gaps that a single
accuracy figure would obscure.

\section{The Benchmark}
\label{sec:dataset}

We evaluate LLM reliability on the \emph{Colombian Legal Reliability Benchmark},
a collection of \num{1042} expert-validated legal questions. The benchmark was
designed to probe legal knowledge and reasoning at increasing levels of
difficulty, following a hybrid methodology that adapts the question typologies
and validation criteria of \citet{dahl2024}, \citet{magesh2025}, and
\citet{arevalo2024} to the Colombian legal order.

\subsection{Question taxonomy}
\label{sec:taxonomy}
The benchmark is organized around a progressive taxonomy of three question
types, each targeting a different level of legal competence and each scored with
a format-appropriate metric (Section~\ref{sec:eval-note}).

\paragraph{Closed (single-best-answer multiple-choice) questions.}
Single-best-answer multiple-choice items with four mutually exclusive options
that probe foundational concepts, definitions, principles, and legal
classifications. Each item has \emph{exactly one} correct option---the model
selects one of the four, not a subset---and a documented legal basis (e.g., a
specific article or ruling). These items are scored by \emph{accuracy}.

\paragraph{Semi-open questions.}
Items requiring a specific, concise answer (up to 150 words) about normative or
conceptual content, without the constraints of multiple choice but preserving
objective evaluation criteria. Adapting the typology of \citet{dahl2024} to
Colombian law, semi-open items are stratified into three levels of complexity
(\emph{low}, \emph{medium}, \emph{high}). \emph{Low}-complexity items cover
easily reproducible knowledge that requires no high-level legal
reasoning---e.g., normative existence, competent authority, legal hierarchy,
identification of the deciding judge, basic definitions, legal classification,
and essential elements of legal institutions. \emph{Medium}-complexity items
require substantive legal understanding that cannot be reproduced verbatim,
demanding some knowledge of the substantive content of a case or norm---e.g.,
the sense of a ruling, literal reproduction of a norm, jurisprudential precedent,
temporal validity, conceptual distinctions, legal requirements and exceptions,
and the orders issued in a decision. \emph{High}-complexity items presuppose
genuine legal-reasoning skills and the ability to synthesize legal information
from unstructured legal texts---e.g., resolving normative conflicts, analyzing
factual backgrounds, adopting a reasoned procedural stance, identifying the
central legal problem and its governing basis, balancing competing principles
and rights, and systematic interpretation.

Within the semi-open type, each complexity level targets a fixed catalogue of
legal sub-tasks (Table~\ref{tab:subtasks}), so that item authors probe a
comparable set of competencies across areas of law.

\begin{table}[htbp]
  \centering
  \small
  \caption{Sub-tasks targeted by semi-open questions at each complexity level,
  adapted from \citet{dahl2024} to the Colombian legal order.}
  \label{tab:subtasks}
  \begin{tabular}{p{0.13\linewidth}p{0.78\linewidth}}
    \toprule
    \textbf{Level} & \textbf{Sub-tasks} \\
    \midrule
    Low & Normative existence; competent authority; deciding judge; legal
    hierarchy; basic definition; legal classification; essential elements. \\
    \addlinespace
    Medium & Sense of the ruling; literal reproduction of a norm; jurisprudential
    precedent; temporal validity; conceptual distinction; legal requirements;
    legal exceptions; impartiality of the judicial order. \\
    \addlinespace
    High & Normative conflict; factual background; procedural stance; legal
    problem; central legal basis; balancing of principles and rights; systematic
    interpretation. \\
    \bottomrule
  \end{tabular}
\end{table}

\paragraph{Open-ended questions.}
Casuistic items that pose a legal problem and whose expert reference answer is
written under the \textbf{IRAC} format---\emph{Issue}, \emph{Rule},
\emph{Application}, and \emph{Conclusion}---synthesizing the full range of legal
competencies. IRAC here describes how the items and their reference answers were
authored; the prompt given to the models does not mention IRAC by name
(Section~\ref{sec:prompting}), and scoring is anchored to the content of the
reference answer rather than to its structure.

Representative example questions for each type are provided in
Appendix~\ref{app:examples}.

\subsection{Legal areas}
\label{sec:areas}
The evaluated benchmark spans ten areas of Colombian law: constitutional,
administrative, criminal, labor, civil, commercial and corporate, procedural,
tax, family, and markets law (competition, consumer protection, data
protection, and intellectual property). Contributors selected the area matching
their expertise, so that items are technically rigorous and representative of
real practice.

\subsection{Dataset composition}
\label{sec:composition}
\begin{figure}[htbp]
  \centering
  \includegraphics[width=\linewidth]{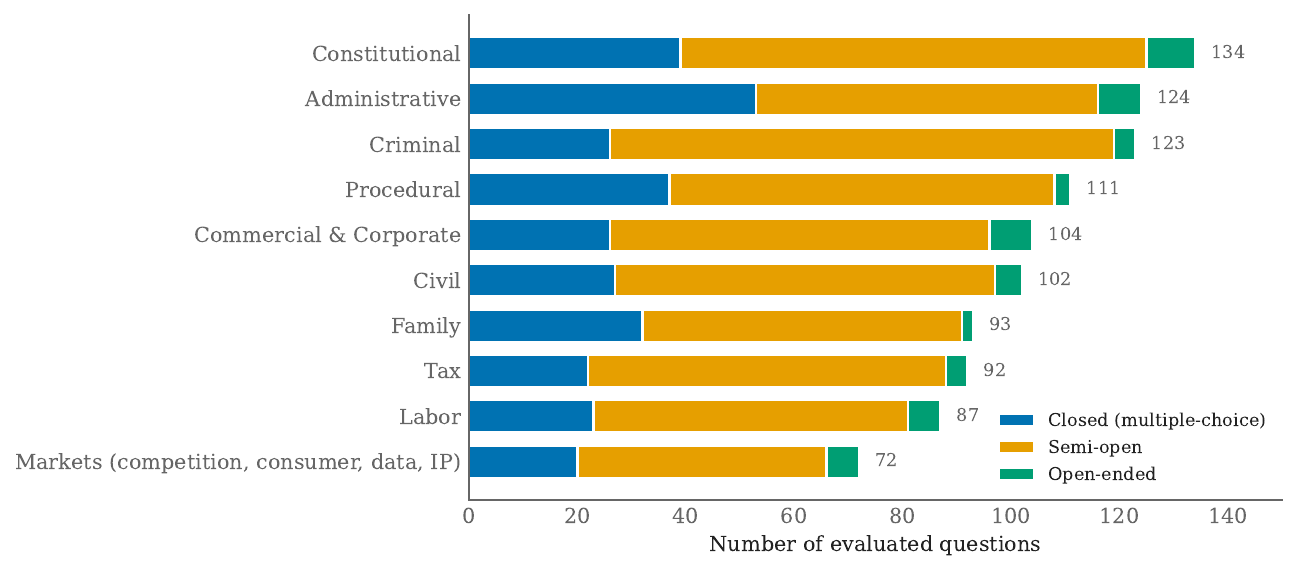}
  \caption{Composition of the evaluated benchmark by legal area and question
  type. Bars are ordered by total number of items; the value at the end of each
  bar is the area total.}
  \label{fig:composition}
\end{figure}

\begin{figure}[htbp]
  \centering
  \includegraphics[width=0.55\linewidth]{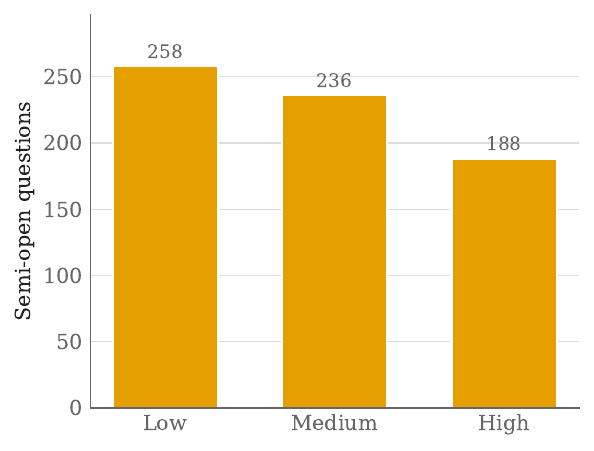}
  \caption{Distribution of semi-open questions across the three complexity
  levels.}
  \label{fig:complexity}
\end{figure}

The evaluated benchmark contains \num{1042} items: \num{305} closed
multiple-choice, \num{682} semi-open, and \num{55} open-ended questions.
Figure~\ref{fig:composition} shows the composition by legal area and question
type, and Table~\ref{tab:composition} reports the exact counts. Semi-open items
are distributed across the three complexity levels as shown in
Figure~\ref{fig:complexity} (\num{258} low, \num{236} medium, \num{188} high).

\begin{table}[htbp]
  \centering
  \small
  \caption{Evaluated benchmark composition by legal area and question type.}
  \label{tab:composition}
  \begin{tabular}{lrrrr}
    \toprule
    \textbf{Legal area} & \textbf{Closed} & \textbf{Semi-open} & \textbf{Open} & \textbf{Total} \\
    \midrule
    Constitutional            & 39 & 86 & 9 & 134 \\
    Administrative            & 53 & 63 & 8 & 124 \\
    Criminal                  & 26 & 93 & 4 & 123 \\
    Procedural                & 37 & 71 & 3 & 111 \\
    Commercial \& corporate   & 26 & 70 & 8 & 104 \\
    Civil                     & 27 & 70 & 5 & 102 \\
    Family                    & 32 & 59 & 2 & 93 \\
    Tax                       & 22 & 66 & 4 & 92 \\
    Labor                     & 23 & 58 & 6 & 87 \\
    Markets (comp./cons./data/IP) & 20 & 46 & 6 & 72 \\
    \midrule
    \textbf{Total}            & \textbf{305} & \textbf{682} & \textbf{55} & \textbf{1042} \\
    \bottomrule
  \end{tabular}
\end{table}

\subsection{A note on evaluation metrics}
\label{sec:eval-note}
Each question type is scored with a format-appropriate metric. Closed
multiple-choice items are scored by \emph{accuracy}. Semi-open and open-ended
items, whose answers are free text, are scored with text-generation and
answer-quality metrics---BERTScore \citep{zhang2020bertscore}, BLEU, ROUGE, and
RAG\-based answer relevancy and correctness \citep{es2024ragas}---together with
a battery of Spanish-adapted readability indices that characterize the style of
the generated answers rather than their correctness. Consequently, different
metrics are computed over different subsets of the benchmark, a distinction we
preserve throughout the analysis.

\section{Dataset Construction Methodology}
\label{sec:methodology}

The benchmark was built through a human-in-the-loop pipeline that combines
student-generated content with multi-stage expert validation, operationalized
on a purpose-built web platform.

\subsection{The Horizontes Conectados platform}
\label{sec:platform}
We designed and deployed \emph{Horizontes Conectados}, a web platform for the
collaborative construction of high-quality legal datasets that integrates
specialized human actors with large language models. The platform defines four
user roles with clearly delimited responsibilities: \emph{students} act as
content generators, producing legal questions in their area of expertise from
established guidelines; a \emph{legal assistant} performs primary review of the
relevance, coherence, and technical quality of each item; \emph{specialized
evaluators} provide expert validation to guarantee legal rigor; and the
\emph{technical team} administers the system and configures the automated
generation processes.

The platform is organized into five functional modules that cover the full
generation-and-evaluation pipeline: (i)~\emph{Questions}, the central
repository that manages each item's life cycle from creation to final
validation; (ii)~\emph{Prompts}, which centralizes and versions the
instructions sent to the models, a determinant of consistency and
comparability across models; (iii)~\emph{Jobs}, the operational core that
orchestrates automated generation and stores the responses produced by each
LLM; (iv)~\emph{Generated Responses}, a query interface for reviewing and
contrasting model outputs; and (v)~a \emph{Dashboard} that consolidates
system-level metrics for operational monitoring.

\subsection{Workshops}
\label{sec:workshops}
\begin{figure}[htbp]
  \centering
  \includegraphics[width=0.62\linewidth]{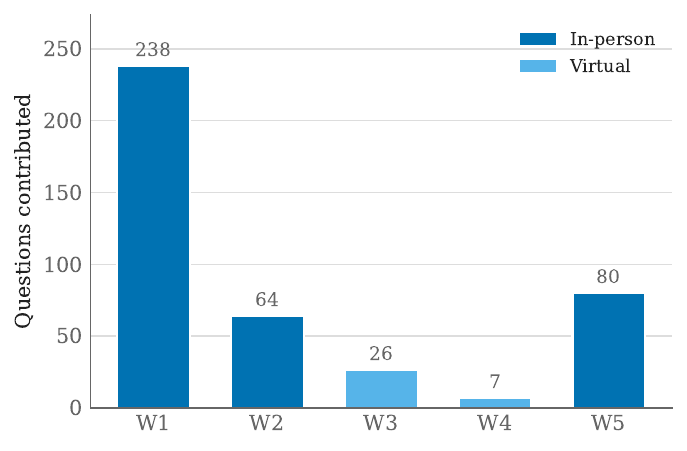}
  \caption{Questions contributed per workshop, colored by modality.}
  \label{fig:workshops}
\end{figure}

The first stage centered on building the legal knowledge base through in-person
and virtual workshops with volunteer undergraduate and graduate students from
the Faculty of Law of Universidad de los Andes. Five workshops were held
between November 2025 and March 2026 (Table~\ref{tab:workshops},
Figure~\ref{fig:workshops}). The two in-person sessions in November 2025
concentrated the largest production of questions.

The workshops were not the only source of items. To broaden coverage across
legal areas and question types, additional questions were authored by the
project's legal team and by students working individually. All items,
regardless of origin, entered the same repository and were subjected to the
identical, multi-stage validation protocol described in Section~\ref{sec:qa}, so
that provenance did not bypass quality control. In total \num{1084} questions
were received---\num{415} from the five workshops and \num{669} from the legal
team and individually working students---of which \num{42} were archived during
validation, yielding the final benchmark of \num{1042} questions.

\begin{table}[htbp]
  \centering
  \small
  \caption{Summary of the question-generation workshops.}
  \label{tab:workshops}
  \begin{tabular}{lllrr}
    \toprule
    \textbf{Workshop} & \textbf{Date} & \textbf{Mode} & \textbf{Questions} & \textbf{Participants} \\
    \midrule
    W1 & 10 Nov 2025 & In-person & 238 & 16 \\
    W2 & 14 Nov 2025 & In-person & 64  & 19 \\
    W3 & 10 Dec 2025 & Virtual   & 26  & 2  \\
    W4 & 15 Dec 2025 & Virtual   & 7   & 2  \\
    W5 & 05 Mar 2026 & In-person & 80  & 10 \\
    \midrule
    \textbf{Total} & & & \textbf{415} & \textbf{49} \\
    \bottomrule
  \end{tabular}
\end{table}

\subsection{Quality-assurance protocol}
\label{sec:qa}
\begin{figure}[htbp]
  \centering
  \includegraphics[width=0.64\linewidth]{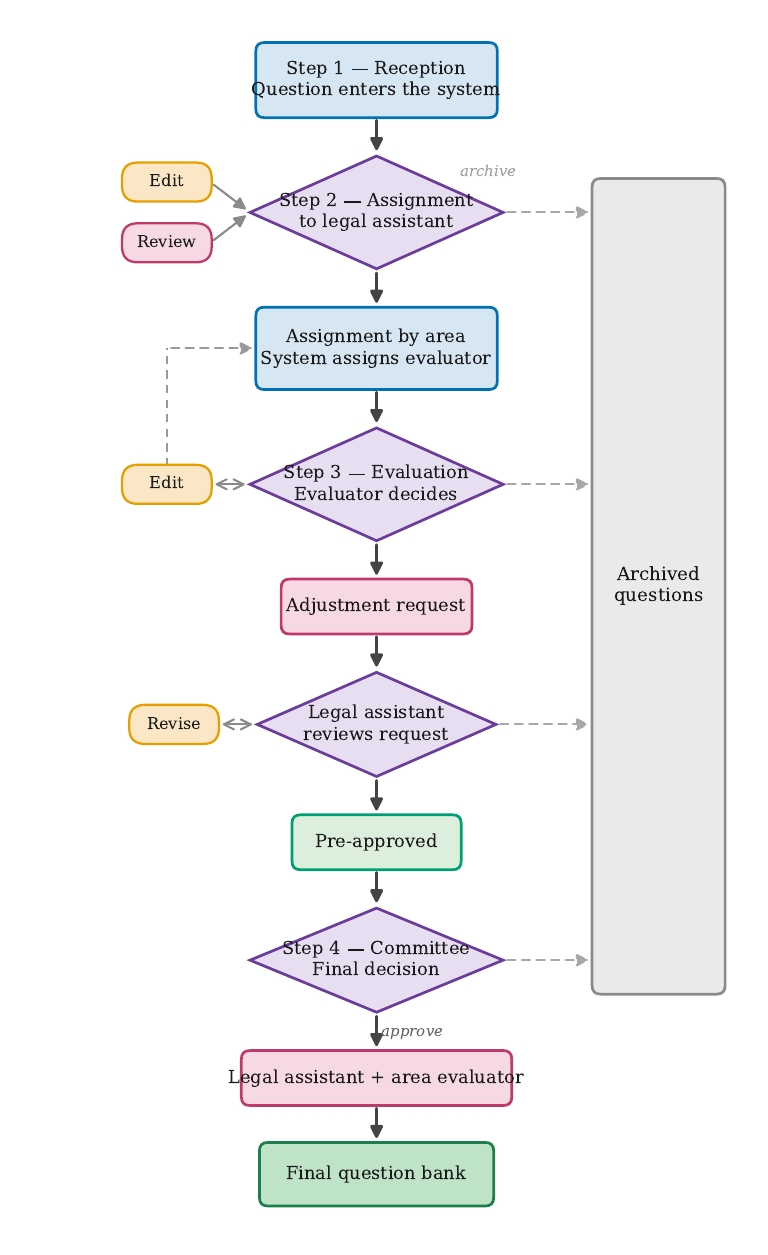}
  \caption{Qualification-and-approval protocol. Every question---whether
  generated in a workshop, by the legal team, or by an individual student---is
  routed through the same four-step protocol (reception, assignment to the legal
  assistant, expert evaluation, and committee decision) before entering the final
  bank; at each decision point an item may be edited, sent back for adjustment,
  or archived.}
  \label{fig:protocol}
\end{figure}

Every collected question was subjected to a structured qualification-and-approval
protocol that guaranteed the quality, relevance, and legal rigor of each item in
the final repository. The protocol (Figure~\ref{fig:protocol}) defines a
sequential route with four steps---reception, assignment to the legal assistant,
expert evaluation, and committee decision---with multiple decision points and
several actors. After an item is received (Step~1) and routed, the substantive
review proceeds in three stages:

\begin{enumerate}
  \item \textbf{Primary review.} Questions generated in the workshops enter the
  system and are automatically assigned to the project's legal assistant, who
  checks the coherence, clarity, and basic relevance of each item. Items needing
  minor corrections are edited and re-submitted; those failing the minimum
  quality criteria are archived and excluded.
  \item \textbf{Expert evaluation.} Items passing the primary review are routed
  to a specialized evaluator by subject area for substantive evaluation. The
  evaluator may (a)~edit the question directly and continue, (b)~request specific
  adjustments from the legal assistant, or (c)~archive the item. Items that pass
  this stage obtain \emph{pre-approved} status.
  \item \textbf{Committee approval.} A committee formed by the legal assistant
  and the area evaluator jointly decides whether each pre-approved item meets the
  required standards. Approved items enter the final bank; the rest are archived.
\end{enumerate}

The protocol thus contemplates three possible trajectories for each item
(Figure~\ref{fig:protocol}): a direct-approval route, an adjustment-and-approval
route, and an archival route. This design subjected the entire collection to a
rigorous, systematic review, ensuring that the final benchmark comprises only
items validated from both a legal and a pedagogical standpoint.

In quantitative terms, the protocol processed the \num{1084} received items and
archived \num{42} of them, so the final bank of \num{1042} reflects a
\num{3.9}\% attrition rate. The adjustment route was heavily used: \num{971} of
the approved items (\num{93}\%) were edited at least once before final approval,
evidence of active curation rather than passive acceptance. The expert-evaluation
stage was staffed by three specialized legal evaluators who covered the ten
areas; most areas were reviewed by a single evaluator, while administrative and
procedural law were each reviewed by two. The protocol did not include
independent double coding with a formal inter-annotator agreement statistic; we
return to this in Section~\ref{sec:limitations}.

\section{Experimental Setup}
\label{sec:setup}

We evaluate \num{15} contemporary large language models on the benchmark of
Section~\ref{sec:dataset}. All models answered the same items under identical,
type-specific prompts, and their responses were scored with metrics matched to
each question format. Every model was queried without internet access or external
tools, relying solely on its parametric knowledge, so that the comparison
reflects what each model has internalized about Colombian law; a web-connected
model that retrieves live sources at answer time would not be comparable and is
therefore out of scope.

\subsection{Evaluated models}
\label{sec:models}
Table~\ref{tab:models} lists the evaluated models, which span both proprietary
and open-weight families from major providers and include general-purpose as
well as reasoning-oriented models. All models were queried through
OpenRouter, a unified inference gateway that provides a consistent
request/response interface across providers. To ensure deterministic and
reproducible outputs, every model was queried with temperature set to
\num{0} and no maximum-token limit, so that answers were never truncated.

\begin{table}[htbp]
  \centering
  \small
  \caption{The \num{15} large language models evaluated on the benchmark.}
  \label{tab:models}
  \begin{tabular}{lll}
    \toprule
    \textbf{Provider} & \textbf{Model} & \textbf{Type} \\
    \midrule
    Amazon      & Nova Lite 1.0                              & General \\
    Anthropic   & Claude Opus 4.5                            & General \\
    Anthropic   & Claude Sonnet 4.5                          & General \\
    Cohere      & Command R (08-2024)                        & General \\
    DeepSeek    & DeepSeek V3.2                              & General \\
    Google      & Gemini 3.1 Pro Preview                     & General \\
    Meta        & Llama 3.3 70B Instruct                     & General \\
    MoonshotAI  & Kimi K2.5                                  & General \\
    NVIDIA      & Llama 3.3 Nemotron Super 49B V1.5          & Reasoning \\
    OpenAI      & GPT-5.4                                    & General \\
    Qwen        & Qwen2.5 72B Instruct                       & General \\
    Qwen        & Qwen3 235B A22B                            & Reasoning \\
    WizardLM    & WizardLM-2 8x22B                           & General \\
    Z.AI        & GLM 4.6                                    & General \\
    Z.AI        & GLM 5.2                                    & General \\
    \bottomrule
  \end{tabular}
\end{table}

\subsection{Prompting strategy}
\label{sec:prompting}
To ensure comparability across models, each question type is answered under a
dedicated, versioned prompt template. Every template casts the model as an
expert in Colombian law, requires it to ground its answer in the applicable
Colombian legal source (statute, article, or jurisprudence), and mandates a
\emph{structured JSON} response so that answers can be parsed and scored
automatically:
\begin{itemize}
  \item \textbf{Closed (multiple-choice).} The model analyzes each of the four
  options under the current Colombian legal order, selects the correct one, and
  returns \texttt{respuesta\_correcta} (the option letter), \texttt{justificacion},
  and \texttt{descarte\_opciones} (a brief reason for discarding each incorrect
  option).
  \item \textbf{Semi-open.} The model produces a concise, structured answer and
  returns \texttt{respuesta}, \texttt{palabras\_clave} (a list of essential
  terms), and \texttt{referencia\_legal} (the cited norm or article).
  \item \textbf{Open-ended.} The model returns a four-part structured legal
  analysis---\texttt{marco\_normativo} (applicable normative framework),
  \texttt{analisis} (reasoning), \texttt{jurisprudencia} (supporting case law),
  and \texttt{conclusion}. The prompt does not name the IRAC method explicitly;
  these fields elicit a response that maps onto the Rule--Application--Conclusion
  structure of the IRAC reference answers, with the Issue set by the question
  itself.
\end{itemize}
Prompt templates are managed and version-controlled in the platform's
\emph{Prompts} module, and the same active version was used for all models,
isolating model capability from prompt variation.

\subsection{Evaluation metrics}
\label{sec:metrics}
Because the three question formats elicit different kinds of answers, they are
scored with different metrics computed over different subsets of the benchmark
(Table~\ref{tab:metrics}).

Because several of these metrics originate in natural-language processing rather
than law, we explain below what each one measures and how to read it. All are
normalized to the $[0,1]$ interval unless noted, with higher values better.

\paragraph{Closed questions.}
Multiple-choice items (\num{305} questions) are scored by \emph{accuracy}, the
proportion of items for which the model selects the single correct option. It is
directly interpretable---an accuracy of \num{0.80} means four of every five
questions were answered correctly---and, because each item has four options,
random guessing yields about \num{0.25}.

\paragraph{Free-text answer quality.}
For semi-open and open-ended items (\num{737} free-text answers), there is no
single ``right string,'' so each model answer is compared against an
expert-written reference answer using four complementary automatic metrics that
capture different notions of similarity:
\begin{itemize}
  \item \textbf{BERTScore} \citep{zhang2020bertscore} measures \emph{semantic}
  similarity---whether the answer \emph{means} the same as the reference---using
  a neural language model, so it credits correct paraphrases that use different
  words. We report its F$_1$ variant.
  \item \textbf{BLEU} and \textbf{ROUGE} measure \emph{lexical overlap}---the
  fraction of words and phrases (\emph{n}-grams) shared with the reference. They
  reward reusing the reference's exact wording; they do \emph{not} assess legal
  correctness, and for free-form legal prose their absolute values are low even
  for good answers. We report ROUGE-1 and ROUGE-L F-scores.
  \item \textbf{RAGAS answer relevancy} \citep{es2024ragas} uses a separate LLM
  as an automated judge to rate how directly the answer \emph{addresses the
  question}---its on-topicness---independently of whether it is correct.
  \item \textbf{RAGAS answer correctness} \citep{es2024ragas} likewise uses an
  LLM judge to rate how \emph{factually consistent} the answer is with the
  expert reference. This is our \emph{primary} measure of free-text quality:
  because the judgment is \emph{anchored to a human-written reference answer}
  rather than being open-ended, LLM-as-a-judge scoring aligns closely with human
  ratings \citep{zheng2023judging,salinas2026lawprofs}. To further strengthen it, we score every
  answer with \emph{two} independent judge models---GLM~5.2 and GPT-4.1---and
  report the mean of their ratings; averaging two judges reduces the idiosyncrasy
  of any single model and yields a more reliable estimate than a single-judge
  score. We use RAGAS v0.4.3 with the default \texttt{answer\_correctness} weighting
  ($0.75$ factual F$_1$ $+\,0.25$ semantic similarity), \texttt{text-embedding-3-large}
  embeddings, and judges queried at temperature~0.
\end{itemize}
BERTScore, BLEU, and ROUGE measure resemblance to the reference rather than legal
validity, and are best read as supporting signals; reference-anchored,
two-judge-averaged RAGAS answer correctness is the metric we rely on for legal
quality. Automatic scoring at this scale is a practical necessity, since
expert human review would require grading the roughly \num{737}$\times$\num{15}$=$\num{11055}
free-text answers produced across all models.

\paragraph{A second, rubric-based judge for validation and error analysis.}
Because a single aggregate score does not reveal \emph{why} an answer fails,
we complement RAGAS with an independent, rubric-based LLM judge applied to a
stratified subset of \num{350} questions (all \num{55} open-ended items plus
\num{295} semi-open items stratified by area and complexity). The judge operates
in two stages: each reference answer is first decomposed \emph{once} into atomic
ideas---each flagged as \emph{core} or not---so that every model is then scored
against the identical set of ideas, making content coverage (\emph{core recall})
comparable across models. For each answer the judge records idea-by-idea coverage
and contradiction, classifies every cited norm as correct, wrong, or
non-existent, tags four non-exclusive error types, and assigns a $1$--$5$
legal-adequacy score. The judged answer is blinded (model identity removed by a
de-fingerprinting pass) and inference uses temperature~0; because the \num{16}
models answer the same questions, confidence intervals use a question-clustered
bootstrap. Crucially, the automatic scores are not assumed to be reliable but checked
against human judgment, because a legal expert team blind-scored a stratified
subset of these judgments, allowing us to quantify how far the automatic metrics
agree with expert assessment and where they diverge
(Section~\ref{sec:res-validation}). This
instrument underpins the citation analysis (Section~\ref{sec:res-citations}) and
the validation of the free-text ranking against both RAGAS and human experts;
RAGAS remains the primary metric reported over the full free-text set.

\paragraph{Answer readability.}
Finally, to characterize the linguistic \emph{style} of the answers---how complex
the prose is, not whether it is correct---we compute five readability indices
adapted to Spanish: Fern\'andez-Huerta, Szigriszt-Pazos, Guti\'errez de Polini,
the $\mu$ index, and Crawford. These estimate how much schooling a reader needs to
understand a text from surface features such as sentence and word length; the
first three are ease scores (higher $=$ easier), while Crawford estimates
years of schooling. We use Spanish-adapted formulas throughout because the
answers are in Spanish. We compute these indices both on the generated answers
and on the expert reference answers, which lets us contrast the stylistic
complexity of model and human prose. These indices are descriptive only and are
not aggregated into the reliability ranking; we analyze their values in
Section~\ref{sec:ext-readability}.

\begin{table}[htbp]
  \centering
  \small
  \caption{Evaluation metrics by question format and the subset each is computed
  over. Arrows indicate whether higher ($\uparrow$) or lower ($\downarrow$)
  values are preferable; readability indices are descriptive (---).}
  \label{tab:metrics}
  \begin{tabular}{lll}
    \toprule
    \textbf{Metric group} & \textbf{Metrics} & \textbf{Subset ($n$)} \\
    \midrule
    Accuracy      & accuracy $\uparrow$ & Closed (305) \\
    Answer quality & BERTScore, BLEU, ROUGE-1/L, & \multirow{2}{*}{Free-text (737)} \\
                   & RAGAS relevancy/correctness $\uparrow$ & \\
    Readability   & Fern\'andez-Huerta, Szigriszt-Pazos, & \multirow{2}{*}{Free-text (737)} \\
    (Spanish)     & Guti\'errez, $\mu$, Crawford (---) & \\
    \bottomrule
  \end{tabular}
\end{table}

\subsection{Evaluation procedure}
\label{sec:procedure}
For each model and question, the platform's \emph{Jobs} module dispatched the
type-specific prompt, collected the structured response, and stored it for
scoring. Answers were then evaluated with the metrics above and aggregated
overall and by legal area and---for semi-open items---by complexity level. The
same item set was presented to every model, so that all reported differences
reflect model behavior rather than differences in the evaluation items.

\section{Results on closed questions}
\label{sec:results}

Table~\ref{tab:mainresults} summarizes the headline metrics for all \num{15}
models across both question families. This section analyzes performance on the
closed multiple-choice questions---overall accuracy and its statistical
reliability, and its structure across legal areas, models, and items
(Sections~\ref{sec:res-overall}--\ref{sec:ext-irt}). Section~\ref{sec:results-gen}
then turns to the free-text, generated answers.

\begin{table}[htbp]
  \centering
  \small
  \caption{Headline results per model, sorted by closed-question accuracy.
  Free-text metrics (BERTScore~F$_1$ and RAGAS relevancy/correctness) are
  computed over the \num{737} open and semi-open items; accuracy over the
  \num{305} closed items. Best value per column in \textbf{bold}.}
  \label{tab:mainresults}
  \begin{tabular}{lcccc}
    \toprule
    & \multicolumn{1}{c}{\textbf{Closed}} & \multicolumn{3}{c}{\textbf{Free-text (open + semi-open)}} \\
    \cmidrule(lr){2-2}\cmidrule(lr){3-5}
    \textbf{Model} & \textbf{Accuracy} & \textbf{BERTScore} & \textbf{RAGAS rel.} & \textbf{RAGAS corr.} \\
    \midrule
    Gemini 3.1 Pro     & \textbf{0.905} & 0.712 & 0.758 & 0.447 \\
    Claude Opus 4.5    & 0.869 & 0.714 & 0.742 & 0.428 \\
    GPT-5.4            & 0.859 & \textbf{0.717} & 0.792 & \textbf{0.451} \\
    GLM 5.2            & 0.856 & 0.708 & 0.745 & 0.389 \\
    Claude Sonnet 4.5  & 0.803 & 0.712 & 0.763 & 0.408 \\
    GLM 4.6            & 0.790 & 0.711 & 0.772 & 0.387 \\
    Kimi K2.5          & 0.754 & 0.703 & 0.732 & 0.354 \\
    DeepSeek V3.2      & 0.728 & 0.707 & 0.785 & 0.359 \\
    WizardLM-2 8x22B   & 0.721 & 0.709 & 0.790 & 0.364 \\
    Qwen3 235B         & 0.708 & 0.712 & 0.785 & 0.342 \\
    Nemotron 49B       & 0.695 & 0.704 & 0.778 & 0.330 \\
    Qwen2.5 72B        & 0.672 & 0.712 & \textbf{0.824} & 0.340 \\
    Llama 3.3 70B      & 0.662 & 0.699 & 0.754 & 0.352 \\
    Nova Lite 1.0      & 0.584 & 0.700 & 0.811 & 0.309 \\
    Command R          & 0.577 & 0.705 & 0.797 & 0.329 \\
    \bottomrule
  \end{tabular}
\end{table}

\subsection{Overall performance on closed questions}
\label{sec:res-overall}
\begin{figure}[htbp]
  \centering
  \includegraphics[width=0.85\linewidth]{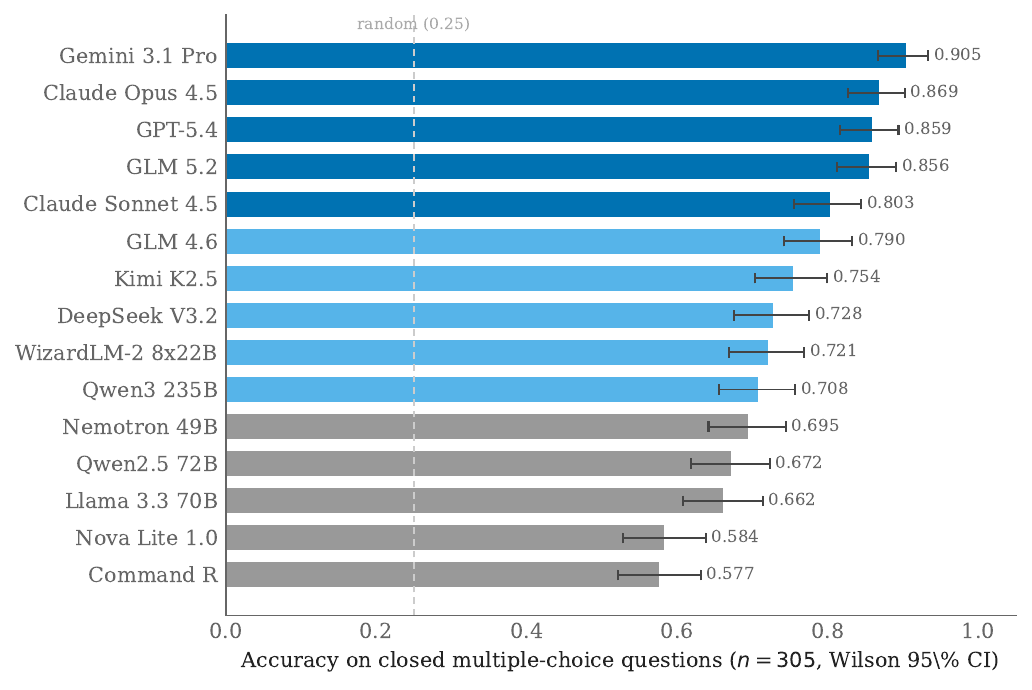}
  \caption{Accuracy on closed multiple-choice questions. Bars are shaded by
  performance tier ($\geq 0.80$, $0.70$--$0.80$, $<0.70$); the dashed line marks
  the random-guessing baseline.}
  \label{fig:accuracy}
\end{figure}

Figure~\ref{fig:accuracy} reports accuracy on the \num{305} closed
multiple-choice items. Performance spans a wide range, from \num{0.905}
(Gemini~3.1~Pro) down to \num{0.577} (Command~R)---all well above the
\num{0.25} random baseline, but with a \num{33}-point gap between the best and
worst models. A leading group of frontier proprietary models (Gemini~3.1~Pro,
Claude~Opus~4.5, GPT-5.4) together with GLM~5.2 exceeds \num{0.85}, while a
second tier clusters between \num{0.75} and \num{0.80}. The lowest-scoring
models---Command~R, Nova~Lite, and the Llama~3.3~70B / Qwen2.5~72B
pair---answer barely more than two thirds of the items correctly, a level of
reliability that is problematic for unsupervised legal use. Error bars show
Wilson \num{95}\% confidence intervals; we test which differences are
statistically reliable in Section~\ref{sec:ext-significance}.

\subsection{Performance across legal areas}
\label{sec:res-area}
\begin{figure}[htbp]
  \centering
  \includegraphics[width=\linewidth]{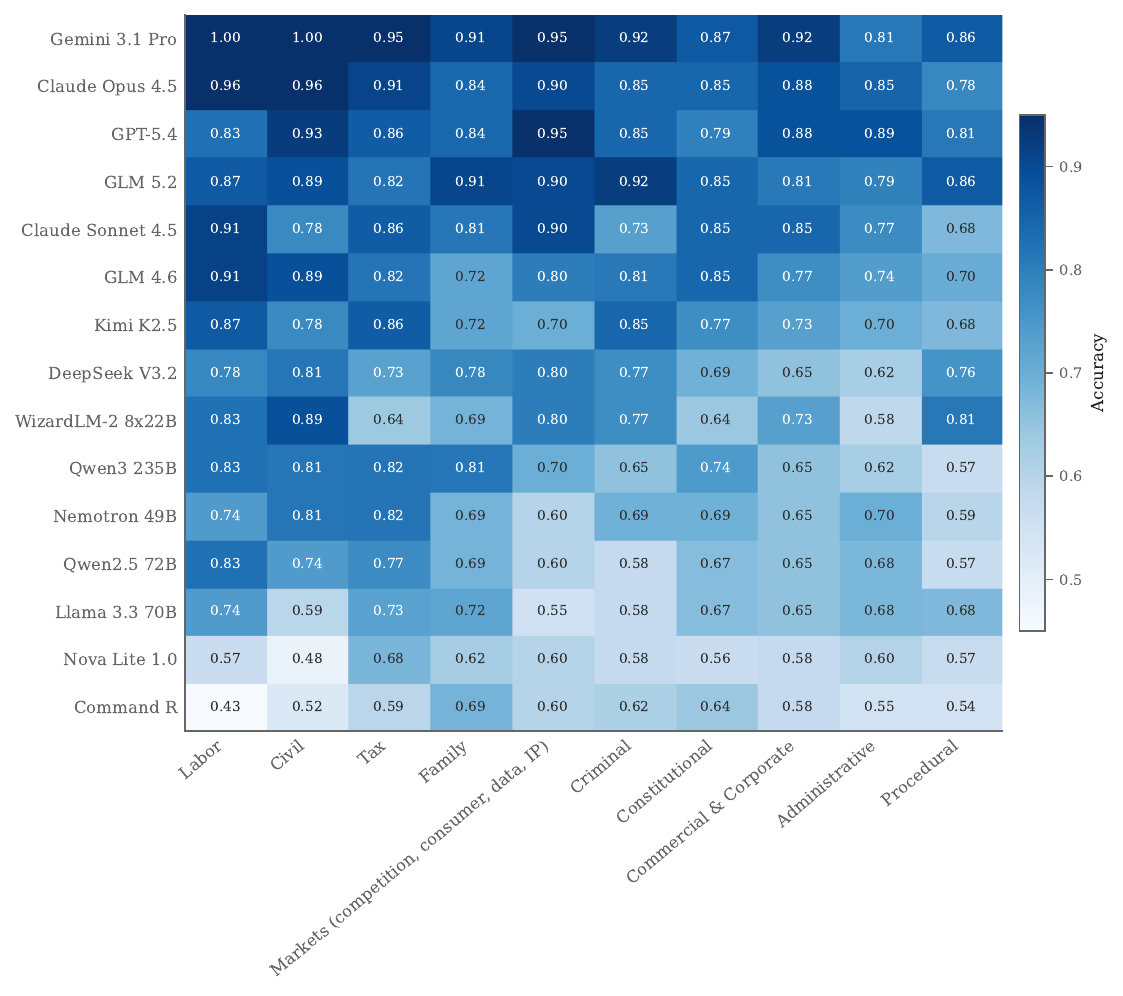}
  \caption{Closed-question accuracy by model (rows, sorted by mean accuracy) and
  legal area (columns, sorted by mean accuracy). Darker cells indicate higher
  accuracy.}
  \label{fig:heatmap}
\end{figure}

Figure~\ref{fig:heatmap} breaks accuracy down by legal area. Two patterns stand
out. First, model ranking is largely stable across areas, with the ordering of
models by mean accuracy preserved within most columns, indicating that stronger
models dominate consistently rather than specializing. Second, model accuracy differs
systematically across areas. Labor, tax, and civil law are answered most
accurately (mean accuracy \num{0.81}, \num{0.80}, \num{0.79}), whereas
procedural and administrative law are where the models perform worst
(mean \num{0.71} each). The
leading models remain robust across areas---Gemini~3.1~Pro and Claude~Opus~4.5
score at or above \num{0.85} in most areas---while weaker models degrade in the
harder areas, approaching chance performance in some cells.

\subsection{Item difficulty and inter-model agreement}
\label{sec:res-agreement}
\begin{figure}[htbp]
  \centering
  \includegraphics[width=0.85\linewidth]{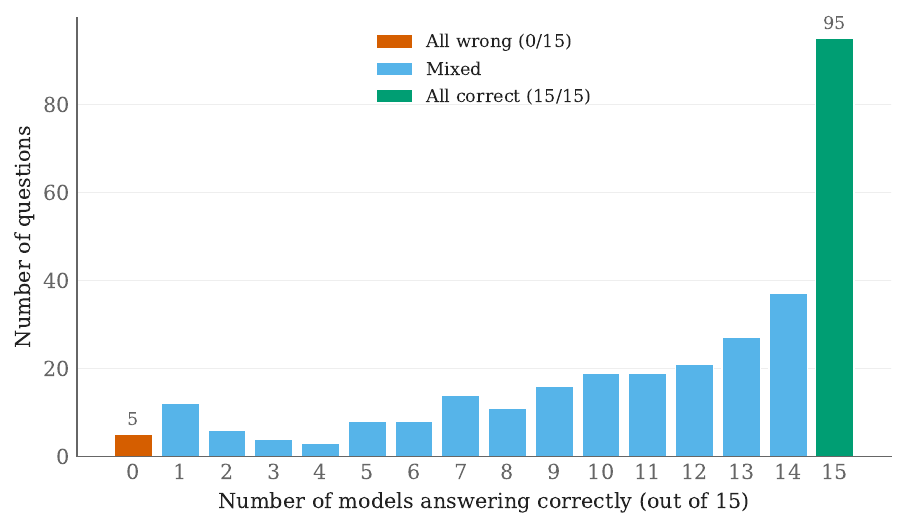}
  \caption{Inter-model agreement on closed questions, showing the number of
  questions for which a given number of models (0--15) answered correctly. A large consensus
  spike at 15/15 contrasts with a short tail of near-universally-failed items.}
  \label{fig:agreement}
\end{figure}

Aggregate scores can hide whether models fail on the \emph{same} questions or on
different ones. To probe this, we count, for each of the \num{305} closed items,
how many of the \num{15} models answer it correctly
(Figure~\ref{fig:agreement}). The distribution is strongly bimodal. At one
extreme, \num{95} items (\num{31.1}\%) are answered correctly by \emph{all}
models---a core of questions that current LLMs handle reliably. At the
other, only \num{5} items (\num{1.6}\%) are missed by \emph{every} model, and a
small tail of \num{22} further items is answered correctly by three or fewer.
The remaining \num{205} items (\num{67.2}\%) are ``mixed,'' splitting the models.
That errors are largely concentrated in mixed items---rather than in
universally-failed ones---indicates that most failures reflect
\emph{model-specific} capability differences rather than intrinsically
unanswerable questions, which makes the choice of model important.

Inspecting the hardest items (Table~\ref{tab:hardest}) helps distinguish genuine
model limitations from items that are simply unfair under our no-context
protocol, since near-universal failure can stem from either cause. Four recurring patterns emerge.
\emph{(i)~Items requiring an external document:} several of the hardest questions
ask the model to reason over a specific attached instrument (e.g., Resolution
No.~368 of 2014) that was not provided in the prompt, so the model cannot answer
regardless of competence. \emph{(ii)~Post-cutoff case law:} questions hinging on
very recent jurisprudence (e.g., Constitutional Court ruling C-517 of 2024) lie
beyond the models' training data. \emph{(iii)~Meta-options:} answers such as
``none of the above'' or ``both'' are systematically missed, as models gravitate
to a concrete plausible option. \emph{(iv)~A flawed item.} Despite the rigorous
multi-stage review, one item was found to contain an internal inconsistency---an
answer option whose spelled-out figure and numeral disagree (it reads ``thirty
percent (10\%)'')---and it is flagged for correction in the released benchmark.
The first two patterns are limitations of the evaluation setup and of parametric
knowledge and motivate retrieval augmentation, whereas this last one is a
benchmark-quality signal that a single item slipped through and should be
revised. By area,
the hardest items concentrate in constitutional, administrative, and procedural
law, mirroring the area-level difficulty of Section~\ref{sec:res-area}.

\begin{table}[htbp]
  \centering
  \small
  \caption{The ten hardest closed questions, by number of models answering
  correctly (out of 15). ``Cause'' classifies the likely reason for
  near-universal failure.}
  \label{tab:hardest}
  \begin{tabular}{clp{0.40\linewidth}l}
    \toprule
    \textbf{Correct} & \textbf{Area} & \textbf{Topic} & \textbf{Likely cause} \\
    \midrule
    0/15 & Procedural     & Consumer-protection competent authority        & Meta-option (``both'') \\
    0/15 & Constitutional & Composition of Congress (upper/lower house)    & Meta-option (``none'') \\
    0/15 & Administrative & Administrative silence (Resolution 368/2014)   & External document \\
    0/15 & Administrative & Disqualifications in state contracts           & Ambiguous \\
    0/15 & Criminal       & Onset of the protected legal interest (life)   & Contested doctrine \\
    1/15 & Commercial     & Who may grant financial leasing                & Nuanced/recent \\
    1/15 & Administrative & Administrative acts (Resolution 368/2014)      & External document \\
    1/15 & Commercial     & Agency problems (case-specific)                & Case-specific \\
    1/15 & Constitutional & Reinforced job stability (ruling C-517/2024)   & Post-cutoff case law \\
    1/15 & Tax            & Penalty for late correction of a return        & Flawed item \\
    \bottomrule
  \end{tabular}
\end{table}

\subsection{Correlated errors and the limits of ensembling}
\label{sec:ext-ensemble}
Section~\ref{sec:res-agreement} showed that errors concentrate in ``mixed''
items. We now ask whether, when models answer a question incorrectly, they tend
to make the \emph{same} mistake. They do, because on items answered incorrectly
the failing models select the \emph{same} distractor \num{87}\% of the time on
average. Errors are therefore
not idiosyncratic noise but shared misconceptions, often driven by an
attractive-but-wrong option.

\begin{figure}[htbp]
  \centering
  \includegraphics[width=0.5\linewidth]{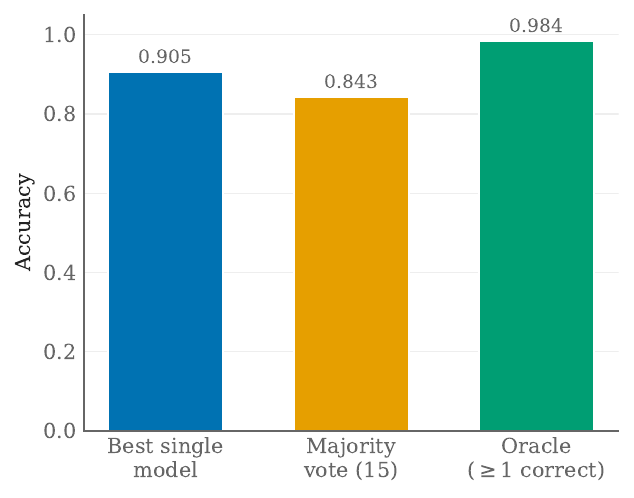}
  \caption{Best single model vs.\ a 15-model majority vote vs.\ the oracle (at
  least one model correct). Majority voting underperforms the best single model
  because model errors are correlated.}
  \label{fig:ensemble}
\end{figure}

This has a direct practical consequence for the common heuristic of polling
several models and taking the majority answer. Because the models fail together,
majority voting does not help. A \num{15}-model majority-vote ensemble reaches
\num{0.843} accuracy, \emph{below} the best single model (\num{0.905}), even
though at least one model answers correctly on \num{98.4}\% of items
(Figure~\ref{fig:ensemble}). The gap between this oracle and the achievable
ensemble is the price of correlated errors. The collective ``knows'' the answer,
but no unsupervised aggregation rule recovers it. Reliability gains must come
from better individual models or from grounding, not from consensus.

\subsection{Model similarity and clustering}
\label{sec:ext-cluster}
\begin{figure}[htbp]
  \centering
  \includegraphics[width=0.72\linewidth]{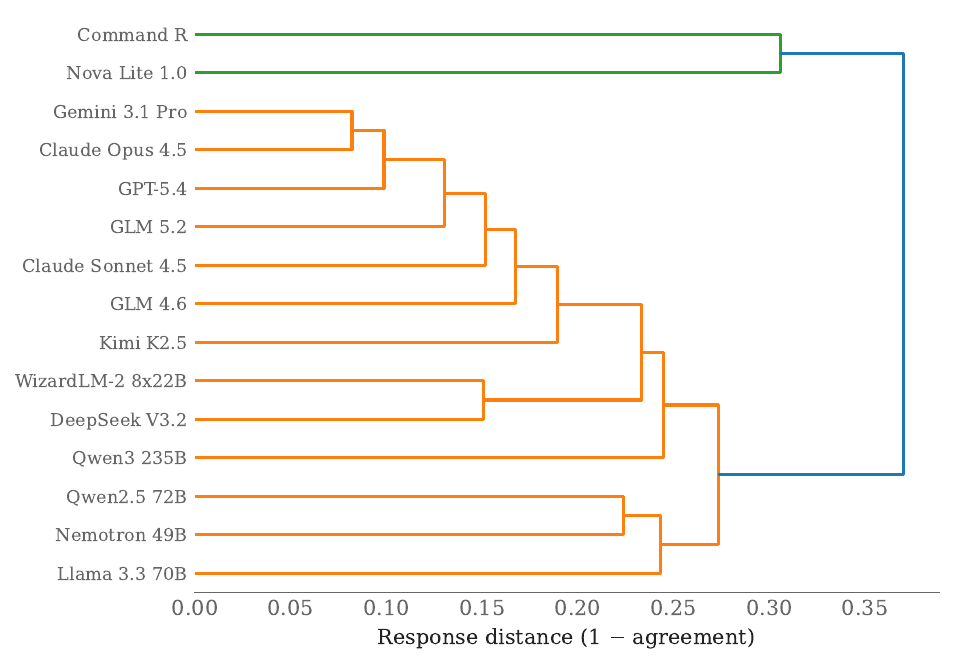}
  \caption{Hierarchical clustering of models by response agreement on closed
  questions (average linkage; distance $=1-$ agreement). Clusters track
  capability tier rather than provider.}
  \label{fig:clustering}
\end{figure}

To see \emph{which} models behave alike, we measure the pairwise agreement
between models (the fraction of items on which they select the same option) and
cluster them hierarchically (Figure~\ref{fig:clustering}). The structure tracks
\emph{capability} rather than vendor. The strongest models (Gemini~3.1~Pro,
Claude~Opus~4.5, GPT-5.4, GLM~5.2) form a tight cluster, while the two weakest
(Command~R, Nova~Lite) join only at the largest distance as an outgroup. Models
from the same provider do not necessarily pair up---Claude~Opus clusters
with Gemini rather than with Claude~Sonnet---indicating that shared answers
reflect a shared competence frontier more than shared training lineage.

\subsection{Statistical significance of model differences}
\label{sec:ext-significance}
\begin{figure}[htbp]
  \centering
  \includegraphics[width=0.72\linewidth]{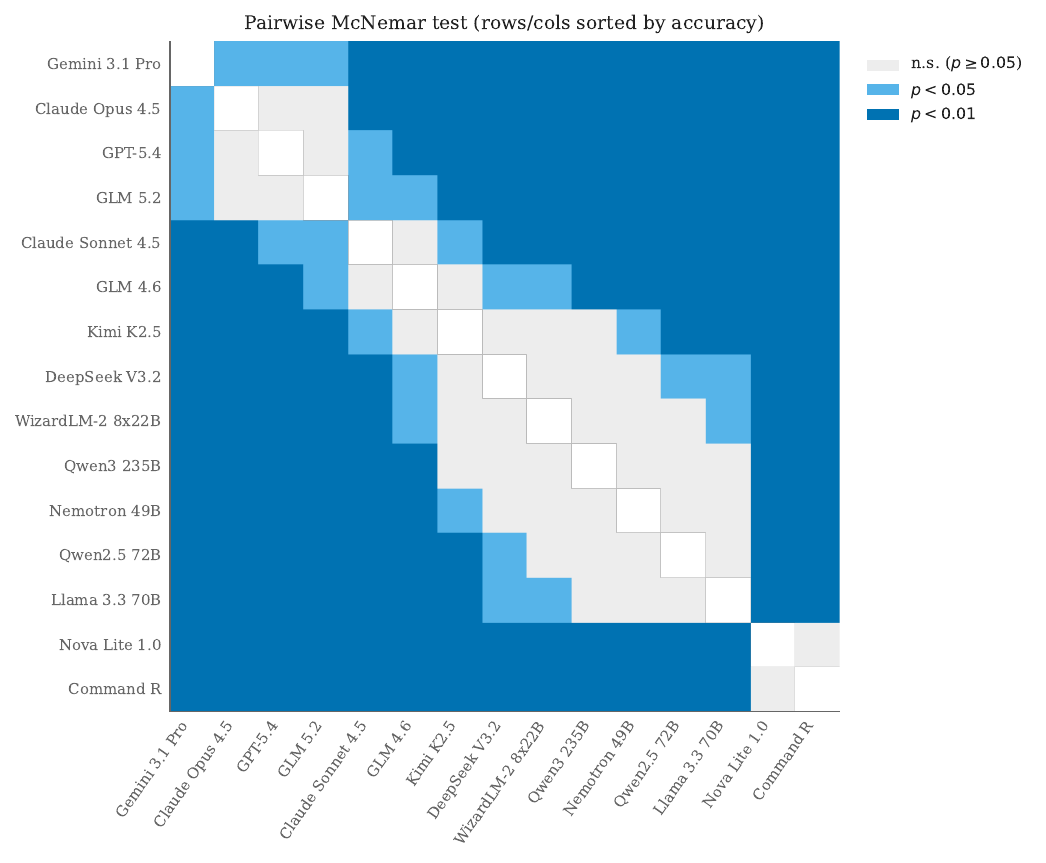}
  \caption{Pairwise McNemar tests on closed-question outcomes (models sorted by
  accuracy). Grey cells mark statistically indistinguishable pairs; the
  block-diagonal grey regions reveal performance tiers.}
  \label{fig:significance}
\end{figure}

With \num{305} items, small accuracy gaps may not be reliable. We add Wilson
\num{95}\% confidence intervals to each model's accuracy
(Figure~\ref{fig:accuracy}) and test all pairwise differences with McNemar's test
on paired item outcomes (Figure~\ref{fig:significance}). Gemini~3.1~Pro is
significantly more accurate than every other model ($p<0.05$). However, the
next tier---Claude~Opus~4.5, GPT-5.4, and GLM~5.2---is statistically
indistinguishable internally, as are several lower tiers; the two weakest models
do not differ significantly from each other. Thus the models form a handful of
statistically distinguishable \emph{tiers} rather than a strict \num{15}-way
ranking, and reported orderings within a tier should not be over-interpreted.

\subsection{The effect of model attributes}
\label{sec:ext-attributes}
\begin{figure}[htbp]
  \centering
  \includegraphics[width=0.55\linewidth]{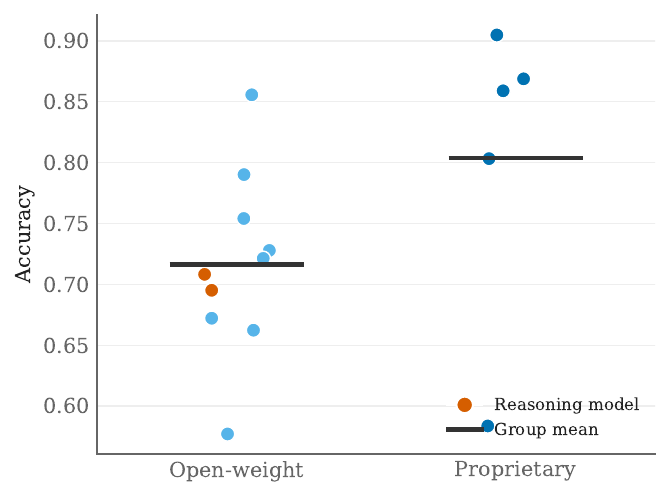}
  \caption{Closed-question accuracy by model access (open-weight vs.\
  proprietary); reasoning-oriented models highlighted. Horizontal bars are group
  means.}
  \label{fig:attributes}
\end{figure}

Two model attributes help explain the spread (Figure~\ref{fig:attributes}).
Proprietary models outperform open-weight ones on average (\num{0.804} vs.\
\num{0.716}; one-sided Mann--Whitney $p=0.038$), though the open-weight group is
heterogeneous and its best members (GLM~5.2, GLM~4.6) rival the proprietary tier.
The two explicitly reasoning-oriented models (Nemotron~49B and Qwen3~235B) also
do \emph{not}
lead. Their mean accuracy (\num{0.702}) is below that of the general-purpose
models (\num{0.752}). On this benchmark---dominated by knowledge recall and
recognition rather than multi-step derivation\footnote{By \emph{recall and
recognition} we mean retrieving or identifying a legal fact the model already
holds (e.g., which article governs a matter, or which of four options is
correct); by \emph{multi-step derivation} we mean chaining several reasoning
steps to work out a conclusion that is not stored as such.}---additional test-time reasoning
offers no advantage, and legal knowledge coverage matters more than reasoning
scaffolding. (With only two reasoning models this comparison is exploratory.)

\subsection{A latent difficulty scale via item response theory}
\label{sec:ext-irt}
\begin{figure}[htbp]
  \centering
  \includegraphics[width=0.65\linewidth]{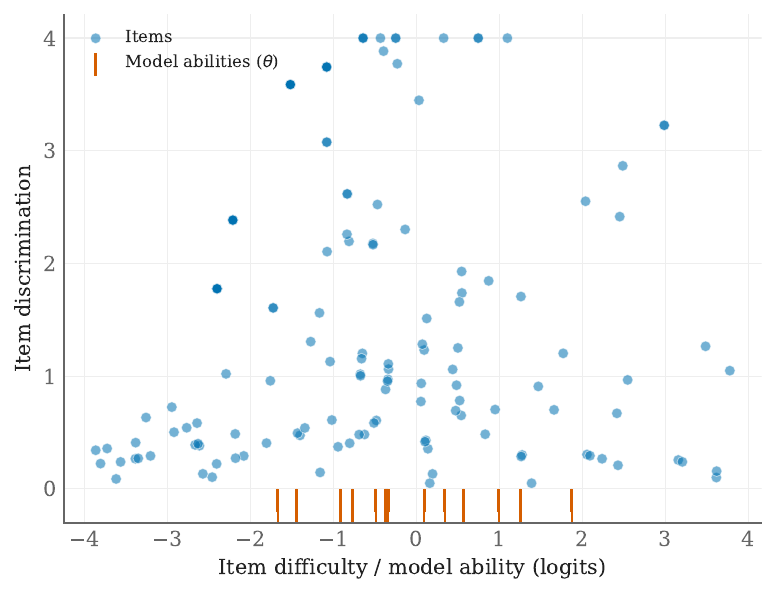}
  \caption{Item response theory (2PL) item map. Each point is a question,
  positioned by estimated difficulty (x) and discrimination (y). Orange ticks
  mark the \num{15} models' estimated abilities on the same logit scale.}
  \label{fig:irt}
\end{figure}

To place items and models on a common scale, we fit a two-parameter logistic
(2PL) item response theory model to the \num{305}$\times$\num{15} matrix of
correct/incorrect outcomes, estimating each item's \emph{difficulty} and
\emph{discrimination} and each model's latent \emph{ability}
(Figure~\ref{fig:irt}). The recovered abilities correlate almost perfectly with
raw accuracy (Spearman $\rho=0.99$), confirming that accuracy is a faithful
summary here; the added value is a calibrated difficulty scale on which model
abilities span the middle of the item range, with a long tail of hard,
high-discrimination items that best separate strong from weak models. Because
each item is answered by only \num{15} models, item-level parameters are
estimated with limited precision and the extremes (items all or none answer
correctly) are only weakly identified; we therefore read the map qualitatively.

\section{Results on free-text generation}
\label{sec:results-gen}
Having characterized recognition on the closed questions, we now turn to
generation. This section covers the quality of the free-text semi-open and
open-ended answers, the reliability of the legal citations they contain, their
validation against human experts, and their readability
(Sections~\ref{sec:res-freetext}--\ref{sec:ext-readability}).

\subsection{Free-text answer quality}
\label{sec:res-freetext}
On the \num{737} open and semi-open items, answer quality is lower and
more compressed than closed-question accuracy suggests. RAGAS answer correctness
does not exceed \num{0.451} for any model (GPT-5.4) on the combined free-text set,
and most models fall between
\num{0.31} and \num{0.45} (Table~\ref{tab:mainresults}). BERTScore is nearly
constant across models (\num{0.699}--\num{0.717}) and thus provides little
discriminative signal, underscoring the value of the RAGAS metrics for this task.

Separating the two free-text formats is instructive. The open-ended (IRAC) items
score \emph{higher} on RAGAS correctness than the semi-open items, and this holds
for \emph{all} \num{15} models (mean \num{0.475} vs.\ \num{0.364}; per-model
difference \num{0.111} on average; paired Wilcoxon $p<0.001$). This is initially
counter-intuitive, as the open-ended items are the most casuistic; the likely
explanation is length. A long, comprehensive IRAC answer gives the
reference-anchored judge more correct content to credit, whereas a semi-open item
is scored against a single precise fact that the model must get exactly right.
Open-ended answers are indeed far longer than semi-open ones (mean \num{308} vs.\
\num{91} words), and answer length correlates positively with correctness both
across models and across individual answers
(Section~\ref{sec:ext-readability}); the open-versus-semi gap should therefore be
read as partly a length effect rather than a purely qualitative one. The aggregate free-text
figure is therefore dominated by---and slightly understated by---the larger
semi-open pool (\num{682} of \num{737} items); we report the two separately where
it matters and treat the open-ended estimate with caution given its smaller size
(\num{55} items).

\begin{figure}[htbp]
  \centering
  \includegraphics[width=\linewidth]{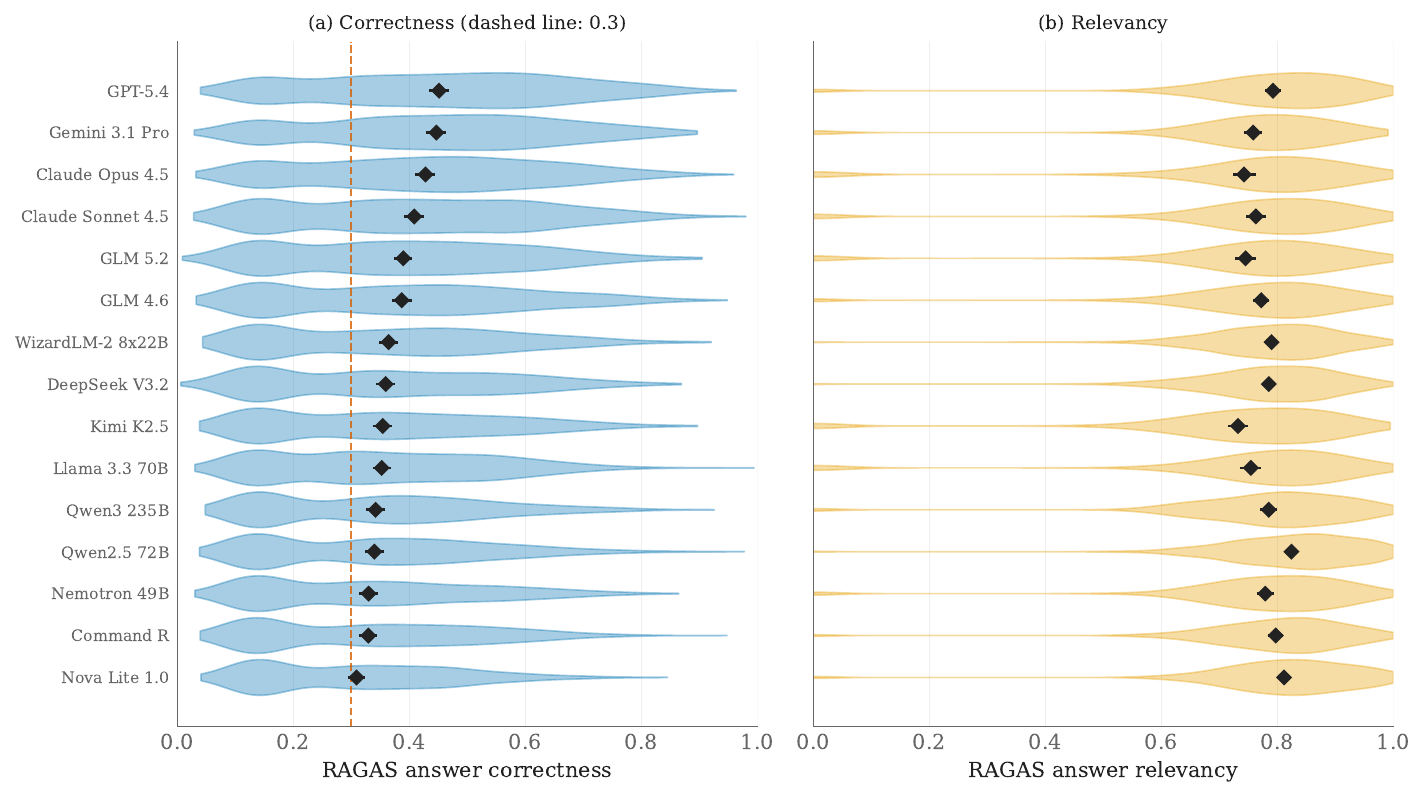}
  \caption{Distribution of RAGAS answer (a)~correctness and (b)~relevancy over
  the \num{737} free-text items, per model (violins; models sorted by mean
  correctness). Diamonds mark the mean and horizontal bars the \num{95}\%
  bootstrap confidence interval. Correctness is low and widely spread for every
  model (dashed line at \num{0.3}), whereas relevancy is uniformly high and
  concentrated.}
  \label{fig:freetext-dist}
\end{figure}

These per-model means, however, conceal a wide and left-skewed spread across
individual answers (Figure~\ref{fig:freetext-dist}). Pooling all \num{11055}
free-text answers, correctness averages \num{0.373} but with a large dispersion
(SD~$=0.199$): \num{39.2}\% of answers score below \num{0.3}, while only
\num{5.9}\% exceed \num{0.7}. The spread is comparably wide \emph{within} every
model (per-model SD \num{0.16}--\num{0.21}), so even the strongest models produce
many low-correctness answers rather than uniformly middling ones. A single mean
therefore overstates the reliability a user should expect from any given answer.

\begin{figure}[htbp]
  \centering
  \includegraphics[width=\linewidth]{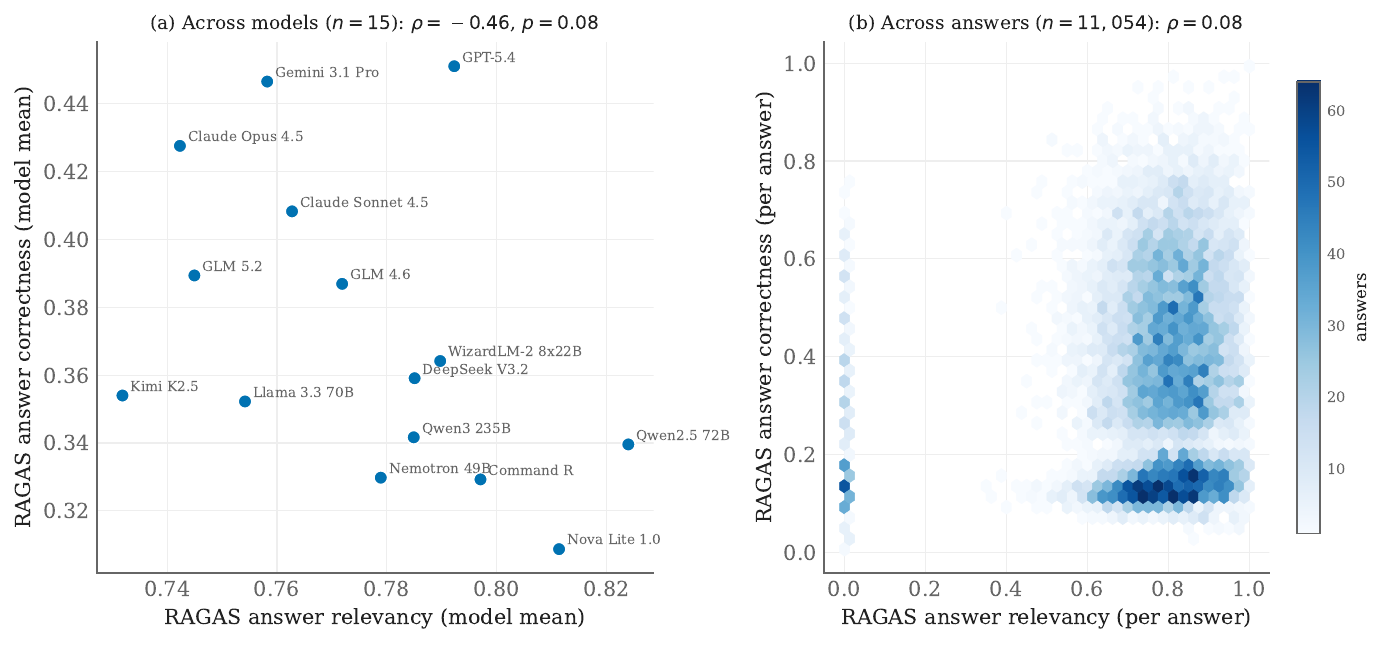}
  \caption{RAGAS answer relevancy versus correctness. (a)~Across the \num{15}
  models (means): higher relevancy does not imply higher correctness
  ($\rho=-0.46$). (b)~Across all \num{11055} individual answers (hex-binned
  density): the two are effectively uncorrelated ($\rho=0.08$). High relevancy
  does not imply correctness at either level.}
  \label{fig:relcorr}
\end{figure}

A dissociation emerges between \emph{relevancy} and \emph{correctness}, and it
holds at two distinct levels of aggregation (Figure~\ref{fig:relcorr}). Relevancy
is uniformly high (\num{0.73}--\num{0.82}). \emph{Across models}, higher mean
relevancy does not track higher mean correctness---if anything it trends the
other way (Spearman $\rho = -0.46$, $p = 0.08$, $n=15$); several of the most
on-topic models, such as Qwen2.5~72B and Nova~Lite, are among the least
factually correct. \emph{Across individual answers}, relevancy is essentially
uninformative about correctness: the answer-level correlation is negligible
($\rho = 0.08$, $n=\num{11055}$), and the mean within-model correlation is just
$\rho = 0.10$ (range $-0.04$ to $0.25$). In other words, a fluent,
on-topic-sounding answer is no more likely to be correct---so relevancy is not a
usable proxy for legal correctness at either the model or the answer level. The
models reliably \emph{sound} responsive while frequently being \emph{wrong}, a
failure mode that makes uncritical use of LLMs in legal settings problematic and
that cautions against relevancy- or coherence-style signals as gates on legal
output.

\begin{figure}[htbp]
  \centering
  \includegraphics[width=0.72\linewidth]{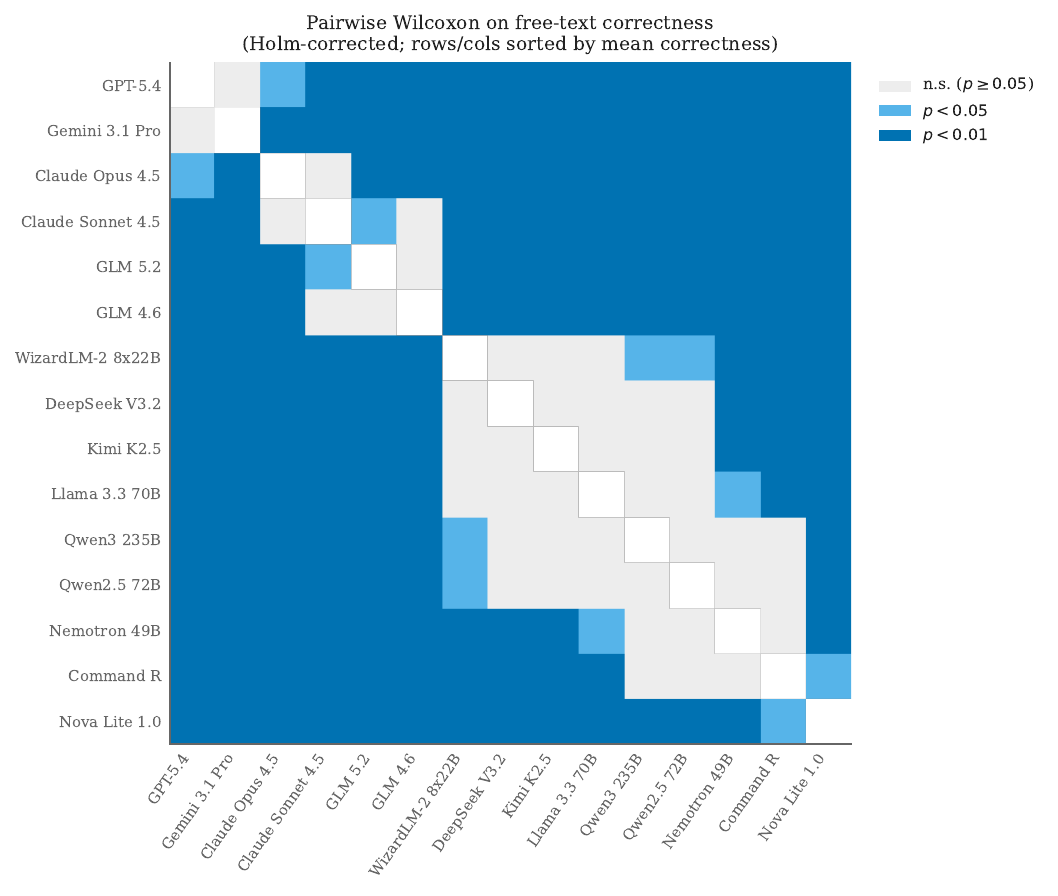}
  \caption{Pairwise paired Wilcoxon signed-rank tests on per-item free-text
  correctness (Holm-corrected; models sorted by mean correctness). Grey cells
  mark statistically indistinguishable pairs; the block structure reveals
  performance tiers.}
  \label{fig:freetext-sig}
\end{figure}

Free-text differences among models are statistically reliable and, as with
closed questions, resolve into tiers rather than a strict ranking. A Friedman
test on per-item correctness across the \num{737} shared items rejects
equality of models ($\chi^2 = 985$, $p < 10^{-100}$), and pairwise paired
Wilcoxon signed-rank tests (Holm-corrected; Figure~\ref{fig:freetext-sig})
localize the structure: GPT-5.4 and Gemini~3.1~Pro lead and are mutually
indistinguishable, Nova~Lite is significantly the weakest, and a broad middle
band of models is largely indistinguishable internally. The \num{95}\% bootstrap
intervals in Figure~\ref{fig:freetext-dist} tell the same story, with overlapping
intervals for adjacent models. As on the closed questions
(Section~\ref{sec:ext-significance}), reported orderings within a tier should not
be over-interpreted.

Performance across the two answer formats is coherent at the
model level. Closed-question accuracy and free-text correctness are strongly
rank-correlated (Spearman $\rho = 0.94$, $p < 0.001$), so a model's standing on
the cheaply scored multiple-choice items is a good predictor of its free-text
reliability.

\paragraph{The hardest free-text items.}
Ranking items by mean correctness across the \num{15} models
(Table~\ref{tab:hardest-freetext})---the free-text analogue of the closed-question
analysis in Section~\ref{sec:res-agreement}---reveals a sharp and telling pattern.
The ten hardest items are all semi-open, and seven of them are pure
\emph{exact-recall} tasks: three ask for the name of the magistrate who authored a
specific ruling (\emph{exact attribution}), and four ask for the exact article of
a specific recent statute---all four from Law 2220 of 2022 on conciliation
(\emph{exact citation}). The remaining three hinge on the holding or reasoning of
a specific recent decision. Strikingly, the single hardest item (mean correctness
\num{0.04}) merely asks who wrote a \num{2025} ruling. As with the closed
questions (Section~\ref{sec:res-agreement}), part of this reflects the no-context,
no-tools protocol and the recency of the cited authorities; but the dominant
signal is that models fail hardest not on legal \emph{reasoning} but on the
\emph{exact facts}---which judge decided, which article governs---that a reliable
legal answer must anchor to, the same brittleness that drives citation
hallucination.

\begin{table}[htbp]
  \centering
  \small
  \caption{The ten hardest free-text items, by mean RAGAS answer correctness
  across the \num{15} models. ``Likely cause'' classifies the dominant reason for
  near-universal failure. All ten are semi-open items.}
  \label{tab:hardest-freetext}
  \begin{tabular}{clp{0.42\linewidth}l}
    \toprule
    \textbf{Corr.} & \textbf{Area} & \textbf{Topic} & \textbf{Likely cause} \\
    \midrule
    0.04 & Constitutional & Reporting magistrate of ruling T-256/2025 (\emph{tutela} vs.\ Meta) & Exact attribution; post-cutoff \\
    0.07 & Constitutional & Deciding judge of ruling T-426/2003 & Exact attribution \\
    0.07 & Constitutional & Reporting magistrate of ruling T-323/2024 (use of AI) & Exact attribution; recent \\
    0.09 & Procedural & Article on the requirements to be a conciliator (Law 2220/2022) & Exact citation; recent statute \\
    0.09 & Administrative & Core holding of ruling SU-167/2023 (direct reparation) & Recent case law; reasoning \\
    0.10 & Tax & Holding of Council of State ruling 26644/2024 & Recent case law \\
    0.10 & Constitutional & What declaring a state of emergency entails (C-145/2020) & Systematic interpretation \\
    0.10 & Procedural & Article on the conciliator's duties (Law 2220/2022) & Exact citation \\
    0.10 & Procedural & Article listing conciliable matters (Law 2220/2022) & Exact citation \\
    0.10 & Procedural & Article on conciliation centers' duties (Law 2220/2022) & Exact citation \\
    \bottomrule
  \end{tabular}
\end{table}

\subsection{Effect of question complexity}
\label{sec:res-complexity}
\begin{figure}[htbp]
  \centering
  \includegraphics[width=0.7\linewidth]{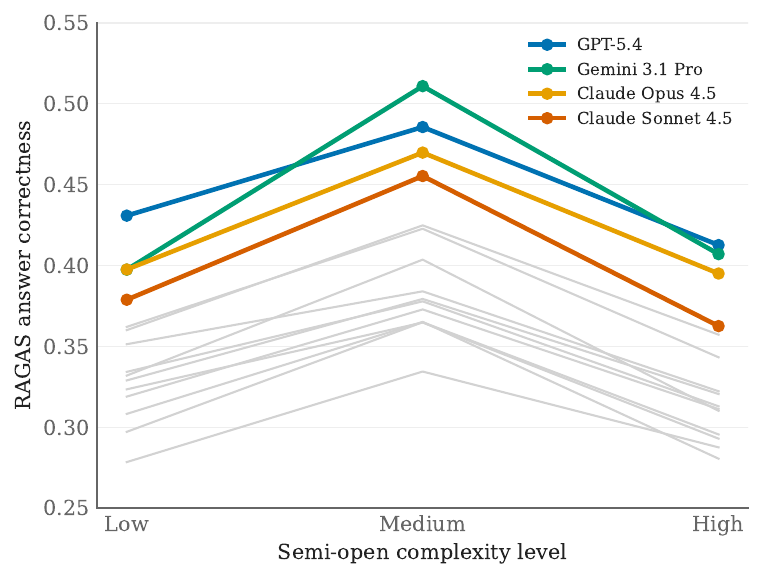}
  \caption{RAGAS answer correctness on semi-open items by complexity level. The
  four highest-scoring models are highlighted; remaining models are shown in
  grey.}
  \label{fig:complexity-res}
\end{figure}

For semi-open items, we stratify correctness by the three complexity levels
(Figure~\ref{fig:complexity-res}). Contrary to a simple monotonic decline,
correctness follows an inverted-U. It peaks at \emph{medium} complexity and is
lowest at \emph{high} complexity for nearly all models. The dip at
\emph{low} complexity is consistent with the nature of those items---normative
existence, exact definitions, and precise citations---which demand verbatim
legal accuracy that models often approximate rather than reproduce. The decline
at \emph{high} complexity reflects the reasoning burden of conflict-of-norms and
systematic-interpretation tasks. The leading models retain a consistent margin
across all three levels.

\begin{figure}[htbp]
  \centering
  \includegraphics[width=0.82\linewidth]{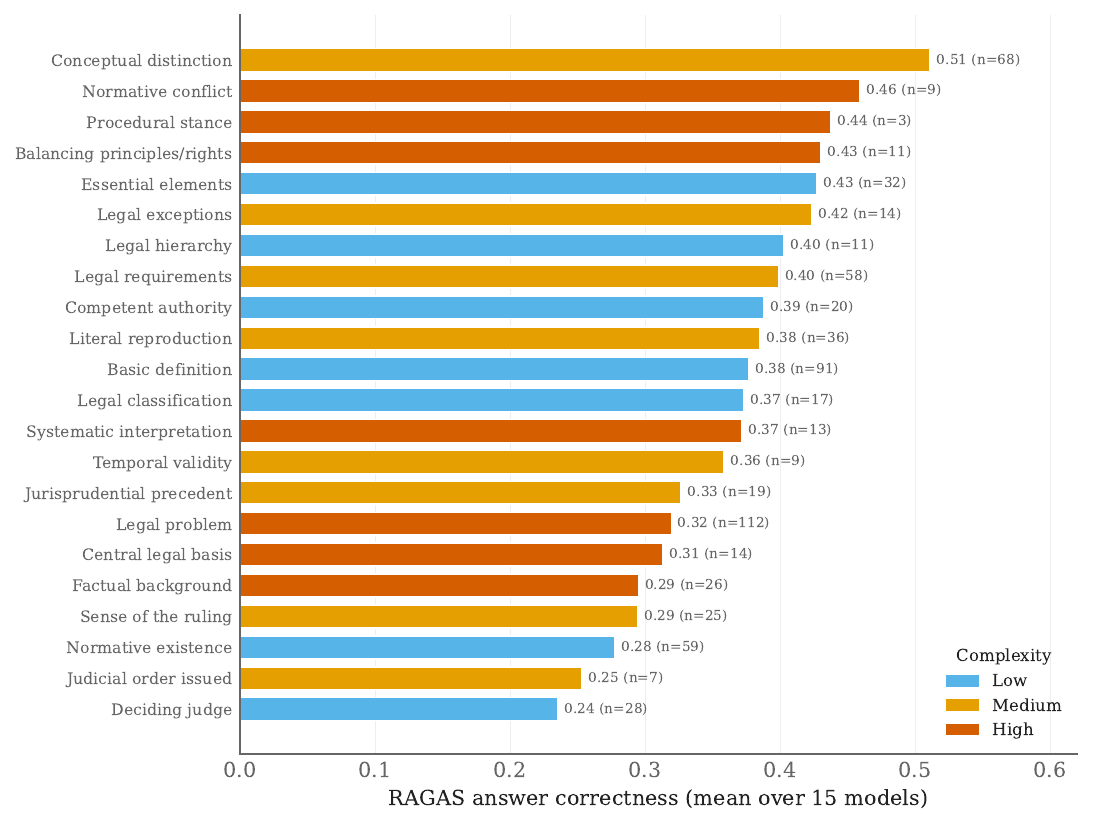}
  \caption{Mean RAGAS correctness by semi-open sub-task (averaged over the
  \num{15} models), colored by complexity level; the item count per sub-task is
  shown at the end of each bar. Low-complexity sub-tasks that hinge on a single
  exact legal fact (deciding judge, normative existence) are among the worst,
  alongside the reasoning-heavy high-complexity sub-tasks.}
  \label{fig:subtask}
\end{figure}

\paragraph{Which sub-tasks are hardest.}
Because each complexity level aggregates finer legal sub-tasks
(Table~\ref{tab:subtasks}), resolving correctness to the sub-task level clarifies
the low-complexity dip (Figure~\ref{fig:subtask}). The lowest-scoring semi-open
sub-tasks are a mix of reasoning-heavy \emph{high}-complexity ones---factual
background (\num{0.29}), central legal basis (\num{0.31}), legal problem
(\num{0.32})---and, tellingly, \emph{low}-complexity ones that turn on a single
exact fact: identifying the deciding judge (\num{0.24}) and normative existence
(\num{0.28}) are the two worst of all. These are precisely the citation-precise
items that require a verbatim fact---which court decided, whether a given norm
exists---rather than legal reasoning, and models approximate them. At the other
end, conceptual distinction (\num{0.51}), legal exceptions (\num{0.42}), and the
balancing of principles (\num{0.43}) score highest. The finest-grained view thus
confirms the mechanism behind the low-complexity dip: brittleness on exact legal
facts, not on reasoning.

\subsection{Free-text quality by legal area}
\label{sec:res-freetext-area}
\begin{figure}[htbp]
  \centering
  \includegraphics[width=\linewidth]{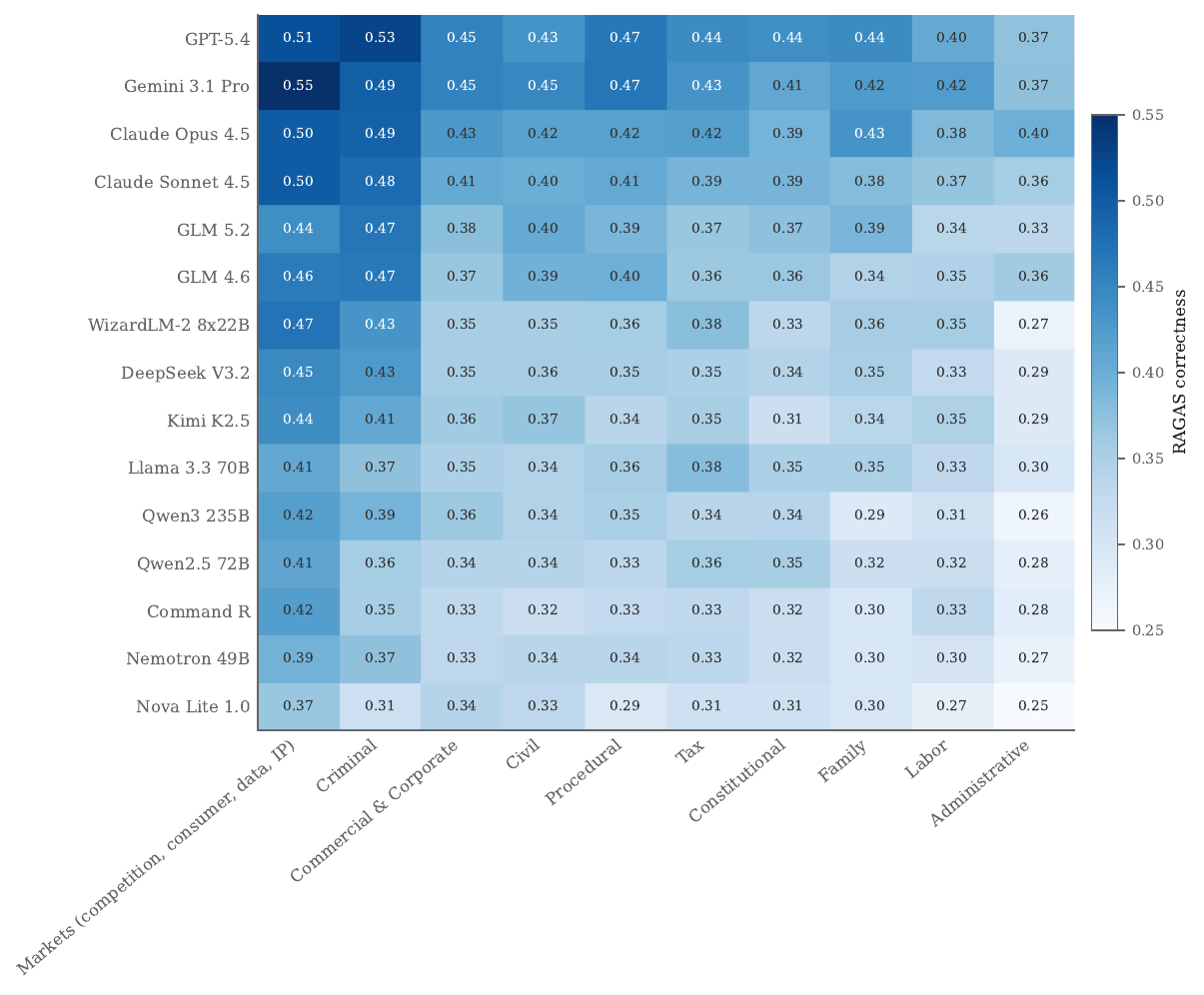}
  \caption{Free-text RAGAS correctness by model (rows, sorted by mean) and legal
  area (columns, sorted by mean). Darker cells indicate higher correctness. The
  analogue of the closed-question accuracy heatmap
  (Figure~\ref{fig:heatmap}).}
  \label{fig:freetext-heatmap}
\end{figure}

Free-text correctness also varies systematically by legal area
(Figure~\ref{fig:freetext-heatmap}), and the pattern only partly matches the
closed-question ranking of Section~\ref{sec:res-area}. Administrative law is
among the hardest in \emph{both} formats---its free-text correctness
(\num{0.31}) is the lowest of any area, and it was also joint-hardest on closed
accuracy. But the correspondence is loose: markets law (\num{0.45}) and criminal
law (\num{0.42}) are the strongest areas for free-text \emph{generation} despite
being only middling on closed \emph{recognition}, whereas procedural
law---joint-hardest on closed questions---sits mid-pack for free text
(\num{0.37}). Recognition and generation therefore stress partly different
competencies across areas. As elsewhere, the leading models (GPT-5.4,
Gemini~3.1~Pro) dominate consistently across areas rather than specializing.

\subsection{Reliability of legal citations}
\label{sec:res-citations}
\begin{figure}[htbp]
  \centering
  \includegraphics[width=0.82\linewidth]{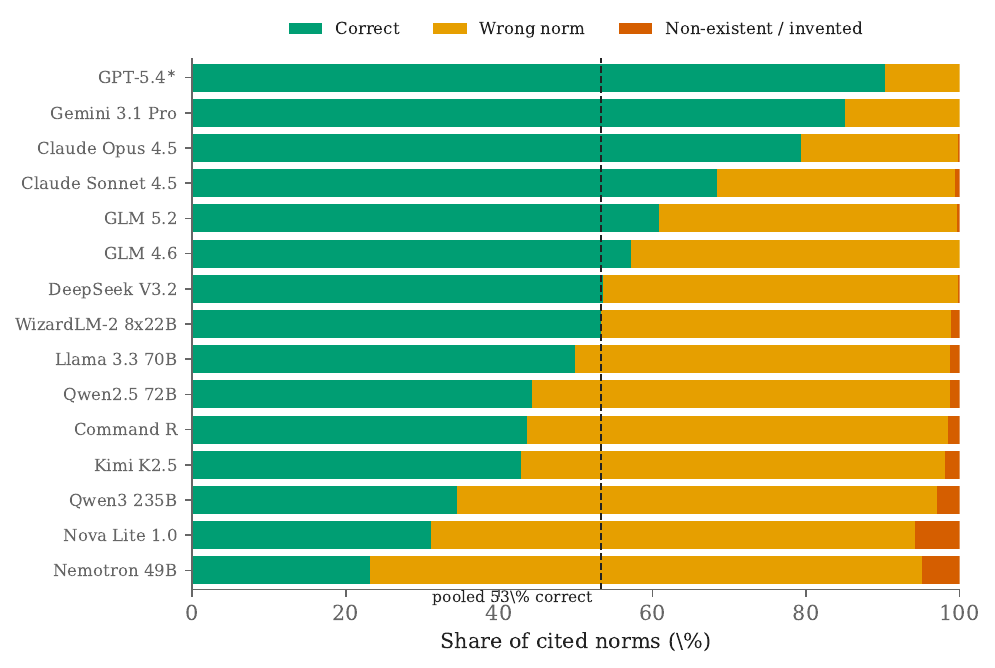}
  \caption{Reliability of normative citations by model (LLM-judge classification
  on the \num{350}-item subset): share of cited norms that are correct, cite the
  wrong norm, or are non-existent. Dashed line marks the pooled correct rate
  (\num{53.2}\%). $^{\ast}$GPT-5.4 is also the judge model
  (Section~\ref{sec:res-validation}).}
  \label{fig:judge-citations}
\end{figure}

A distinctive risk of legal LLMs is not omission but \emph{fabrication}: citing a
norm that does not say what the model claims, or that does not exist
\citep{dahl2024,magesh2025}. The rubric-based judge (Section~\ref{sec:metrics})
classifies every cited norm carrying a numeric identifier as \emph{correct},
\emph{wrong} (a real norm that does not support the claim), or
\emph{non-existent}. Across the \num{15} models on the \num{350}-item subset
(\num{5140} such citations), only \num{53.2}\% are correct: \num{45.2}\% point to
the wrong norm and \num{1.6}\% are non-existent
(Figure~\ref{fig:judge-citations}). Citation accuracy separates models sharply,
from \num{90.3}\% (GPT-5.4) down to \num{23.1}\% (Nemotron~49B), and the
fabrication of non-existent norms, though rare on average, concentrates in the
weakest systems (Nova~Lite \num{5.8}\%, Nemotron \num{4.9}\%).

This is the single most legally consequential finding, and it is also the
best-supported one, because the human validation below flags it directly. On the subset of
answers that cite a norm, the judge's ``wrong-or-invented citation'' flag agrees
with the three-person legal team---who verified each cited norm individually---in
\num{91}\% of cases, with Gwet's AC1 of \num{0.88} and \num{100}\% specificity
(Section~\ref{sec:res-validation}). When the judge reports that a citation is
wrong or invented, the human verification concurs.

\subsection{Convergent validity and human validation}
\label{sec:res-validation}
Because the free-text conclusions rest on automatic judges, we test how far they
survive independent scrutiny---from a second, methodologically different
automatic instrument and from human experts.

\begin{figure}[htbp]
  \centering
  \includegraphics[width=0.6\linewidth]{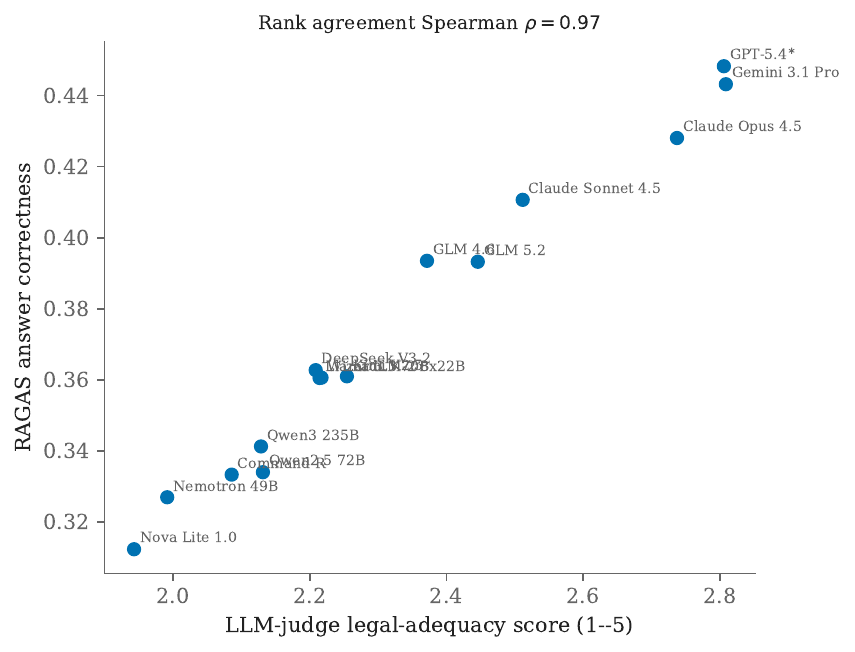}
  \caption{Convergent validity: each model's mean rubric-based LLM-judge score
  against its mean RAGAS answer correctness (\num{15} models, \num{350}-item
  subset). Two methodologically independent instruments agree closely on the
  ranking. $^{\ast}$GPT-5.4 is also the judge model.}
  \label{fig:judge-ragas}
\end{figure}

\paragraph{Convergent validity with RAGAS.}
The rubric-based judge and RAGAS are built differently---a decomposed
idea-coverage rubric versus a reference-anchored factual-consistency score---yet
they rank the \num{15} models almost identically: Spearman $\rho=\num{0.97}$,
Kendall $\tau=\num{0.89}$ (Figure~\ref{fig:judge-ragas}). At the level of
individual answers they correlate only moderately (Spearman $\rho=\num{0.56}$).
This pattern---strong agreement on the ordering, looser agreement
answer-by-answer---is the signature of two valid measures carrying independent
noise, and it means the free-text ranking does not depend on the idiosyncrasies
of any one scoring method.

\begin{figure}[htbp]
  \centering
  \includegraphics[width=0.62\linewidth]{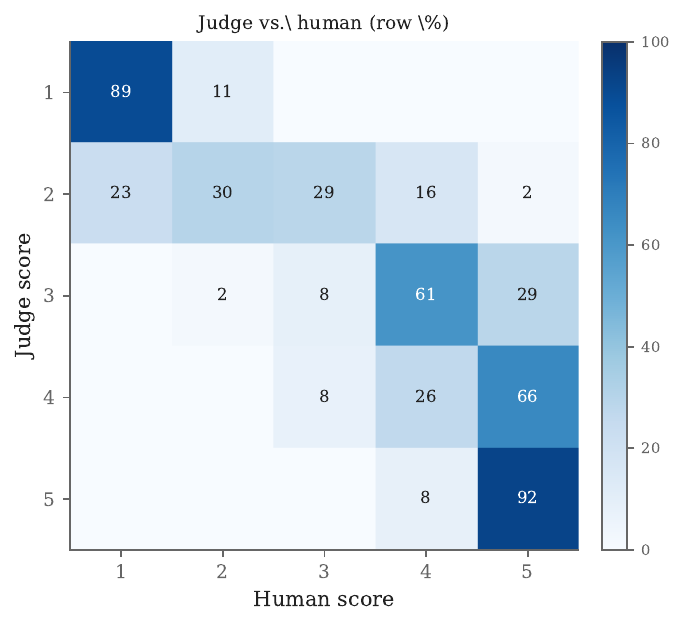}
  \caption{Human validation of the LLM judge: confusion matrix of judge versus
  expert-team scores on the $1$--$5$ scale (row-normalized, \num{323} blind
  judgments on the \num{15}-model subset). The extremes are reliable; the
  mid-scale is shifted down by about one point, a monotone shift that preserves
  the model ranking.}
  \label{fig:human-confusion}
\end{figure}

\paragraph{Human validation.}
A legal expert, supported by two students who verified each cited norm
individually, blind-scored \num{323} judge decisions on the \num{15}-model
subset, seeing exactly the (de-fingerprinted) answer text the judge saw and
neither its verdict nor the model's identity. Agreement on the $1$--$5$ scale is
at the accepted threshold---quadratic Cohen's $\kappa=\num{0.635}$, ordinal
Krippendorff's $\alpha=\num{0.667}$, \num{83}\% of cases within one point---and,
crucially, the judge orders the models almost as the humans do (Spearman
$\rho=\num{0.88}$; Figure~\ref{fig:human-confusion}). The judge is, however,
systematically \emph{stricter} than the human team by about half a point (mean
difference $\num{-0.53}$, CI excludes zero). The consequence is clear-cut. The
\emph{comparative} claims---the ranking, the significant pairwise differences,
the agreement with RAGAS---hold, whereas \emph{absolute} levels from the judge
read low. Reported side by side, the mean legal-adequacy score is \num{2.32}/5 for the
judge versus \num{2.85}/5 for the human team, and the share of answers deemed
useful to a practising lawyer is \num{8.8}\% (judge) versus \num{41.8}\% (human);
the judge's usefulness rate is best read as a conservative floor, since it misses
roughly four of five answers the humans found useful. The confusion matrix shows
why the ranking nonetheless survives: the extremes are reliable (a judge score of
\num{1} is human \num{1} \num{89}\% of the time, a judge \num{5} is human \num{5}
\num{90}\% of the time) and the mid-scale shift is monotone.

\paragraph{Judge self-preference.}
GPT-5.4 is both the rubric judge and one of the evaluated models, which invites a
self-preference concern \citep{zheng2023judging}. The human validation measures
it directly, since the judge's bias against GPT-5.4 is $\num{-0.70}$ points, more
severe than its mean bias of $\num{-0.48}$ across the other models (range
$[\num{-1.04},\num{+0.36}]$)---that is, the judge is \emph{harder} on its own
family, not softer, so there is no evidence of self-preference. We therefore
retain GPT-5.4 in both roles (removing it would itself bias the comparison) and
mark it in the relevant figures. The independent RAGAS ranking and the human
ranking, both of which agree with the judge, further insulate the conclusions
from any single-judge effect.

\subsection{Answer length and readability}
\label{sec:ext-readability}
Having the raw answer text lets us measure a stylistic variable the format
analysis left implicit---\emph{answer length}---which turns out to be the
operative variable behind the Spanish readability signal.

\begin{figure}[htbp]
  \centering
  \includegraphics[width=\linewidth]{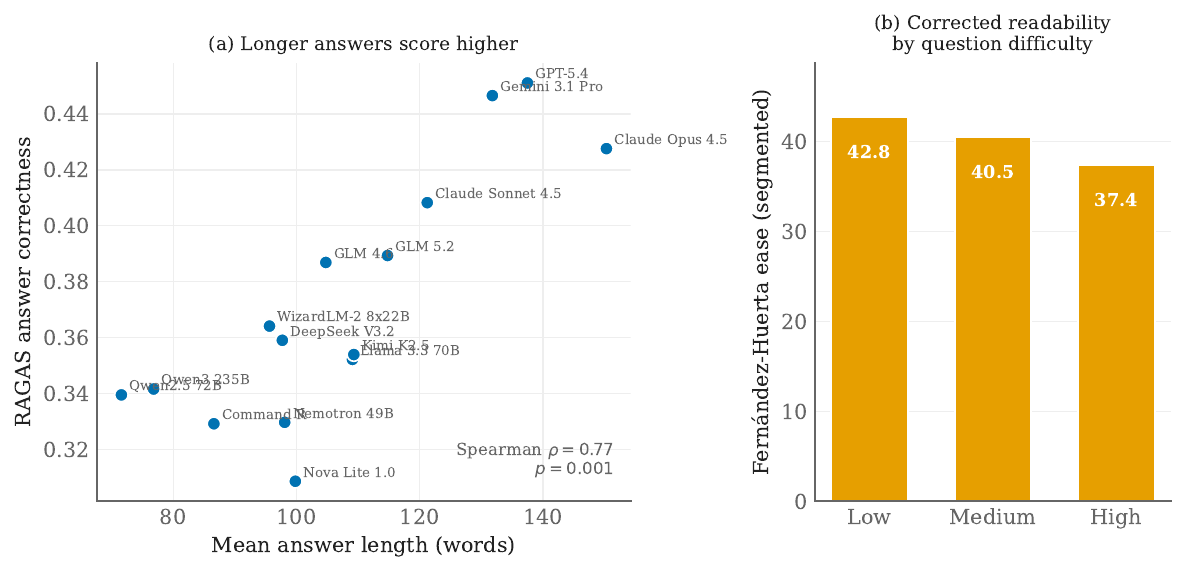}
  \caption{Answer length and readability. (a)~Mean answer length (words) vs.\
  RAGAS correctness across models: longer answers score higher
  ($\rho=+0.78$). (b)~Fern\'andez-Huerta ease by semi-open complexity level;
  values sit inside the calibrated $0$--$100$ band. Higher ease $=$ easier to
  read.}
  \label{fig:readability}
\end{figure}

\paragraph{Length is the variable behind the density signal.}
Generated answers vary more than twofold in length, from a mean of \num{72}
words (Qwen2.5~72B) to \num{150} (Claude~Opus~4.5), with the frontier models the
most verbose (GPT-5.4 \num{137}, Gemini~3.1~Pro \num{132}). Length tracks
correctness: across models, mean answer length correlates with RAGAS correctness
at Spearman $\rho=+0.78$ ($p<0.001$; Figure~\ref{fig:readability}a), and the
relationship persists at the level of individual answers ($\rho=+0.21$,
$n=\num{11055}$, $p<10^{-100}$). What might look like ``denser prose is more
correct'' is a length effect: because the Spanish ease indices are
dominated by sentence length, controlling for length removes the apparent
density--correctness link (partial Spearman $\rho=-0.31$, $p=0.27$ across
models). The more robust reading is straightforward---stronger models write
\emph{more}, and RAGAS, an F1 over reference-anchored claims, credits the
additional correct content they include. Length is therefore a confounder to
bear in mind for the open-versus-semi-open gap of
Section~\ref{sec:res-freetext} (open answers average \num{308} words against
\num{91} for semi-open, and are longer \emph{and} more correct for all \num{15}
models) and for between-model comparisons.

\paragraph{Readability.}
We compute the five Spanish ease indices with sentence
segmentation. The
generated answers sit \emph{within} the ordinary readability band: mean
Fern\'andez-Huerta ease is \num{40.3} across models
(Figure~\ref{fig:readability}b), consistent with dense but well-formed legal
prose. Complexity still rises modestly with question difficulty: for semi-open
items mean ease declines from \num{42.8} at low complexity to \num{40.5} at
medium and \num{37.4} at high. The $\mu$ and Crawford indices remain the least
informative (near-zero association with correctness).

\begin{figure}[htbp]
  \centering
  \includegraphics[width=\linewidth]{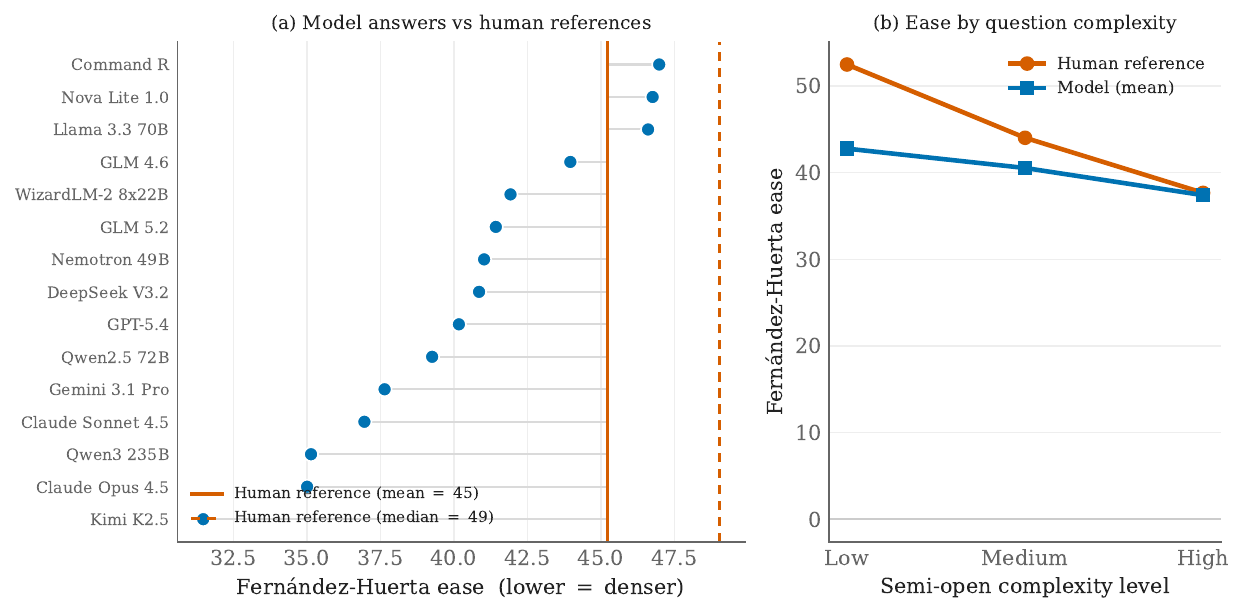}
  \caption{Readability of model answers versus the human expert
  references (Fern\'andez-Huerta ease; higher $=$
  easier). (a)~Each model's mean ease (points) against the human-reference mean
  (solid) and median (dashed): most models write somewhat denser prose than the
  human references, but within the ordinary range. (b)~Ease by semi-open
  complexity level: the human--model gap is largest at low complexity and closes
  at high complexity.}
  \label{fig:readability-human}
\end{figure}

\paragraph{Model answers versus human references.}
Computing the same indices on the expert-written reference answers, models
still write somewhat denser prose than the human experts, but the gap is modest
rather than dramatic. Mean Fern\'andez-Huerta ease is \num{45.2} for the
references (median \num{49.0}) versus \num{40.3} across models; \num{12} of
\num{15} models fall below the human mean and all \num{15} below the human median
(one-sided Wilcoxon $p<10^{-17}$; Figure~\ref{fig:readability-human}a). The same
ordering holds on Szigriszt-Pazos (\num{40.8} vs.\ \num{35.7}) and
Guti\'errez de Polini (\num{36.5} vs.\ \num{33.2}). The divergence is largest at
\emph{low} complexity---where a human expert answers a simple question plainly
($\text{ease}=\num{52.5}$) but the model still produces fuller, more technical
prose (\num{42.8})---and essentially closes at \emph{high} complexity
(\num{37.7} vs.\ \num{37.4}; Figure~\ref{fig:readability-human}b). The densest
answers do not come simply from the strongest models: the lowest-ease answers
are written by a mix of strong and
mid-tier systems (Kimi~K2.5, Claude~Opus~4.5, Qwen3~235B), so density is best
read as a stylistic trait only weakly related to quality once length is
accounted for. In short, models tend to over-elaborate relative to expert
references---most visibly on easy questions---while writing at a broadly
comparable reading level.

\section{Discussion}
\label{sec:discussion}

\subsection{Reliability remains below the threshold for unsupervised legal use}
Across the evaluation, the results caution against deploying LLMs autonomously to
answer questions about the Colombian legal system. While the strongest models answer
closed questions accurately (Gemini~3.1~Pro at \num{0.905};
Figure~\ref{fig:accuracy}), free-text answer correctness tells a different story:
no model exceeds \num{0.451} RAGAS correctness, and most sit between \num{0.31}
and \num{0.45} (Table~\ref{tab:mainresults}). This gap between recognizing the
correct option and \emph{producing} a correct legal answer is central to the
reliability problem. Our findings extend to the Colombian, civil-law setting the
cautionary evidence documented for U.S. common-law systems, where general-purpose
models hallucinate on a majority of legal queries and even
retrieval-augmented tools remain far from hallucination-free
\citep{dahl2024,magesh2025}. They also update the earlier Colombian assessment of
\citet{arevalo2024}. The newest frontier models are more capable on
structured questions than the models evaluated in 2024, yet on open legal
analysis they still fall short of a standard acceptable for practice.

This conclusion does not rest on a single automatic metric. The free-text ranking
is corroborated by three methodologically independent instruments---RAGAS, a
rubric-based LLM judge, and blind human expert scoring---which agree closely on
the ordering of models (pairwise rank correlations $\rho\geq\num{0.88}$), even as
the automatic judges read about half a point low in absolute terms
(Section~\ref{sec:res-validation}). The most legally consequential failure this
triangulation exposes is citation reliability. Across models, only \num{53.2}\% of
the norms the models cite are correct, \num{45.2}\% cite the wrong norm, and
\num{1.6}\% are non-existent---a result the human team confirmed norm by norm
(Section~\ref{sec:res-citations}). For a legal user, a wrong or invented citation
is more dangerous than a hedged non-answer, because it is both actionable and
plausible; this is the concrete form the reliability gap takes in practice, and it
echoes the citation-hallucination findings that motivated \citet{dahl2024} and
\citet{magesh2025}.

\subsection{Fluency does not imply correctness}
The most actionable finding for practitioners is the dissociation between
answer relevancy and correctness (Figure~\ref{fig:relcorr}). Every model produces
highly on-topic answers (relevancy \num{0.73}--\num{0.82}), but relevancy does
not track factual correctness at either level of aggregation. Across models the
two even trend apart ($\rho = -0.46$; some of the most relevant-sounding models,
Qwen2.5~72B and Nova~Lite, are among the least correct), and across individual
answers the correlation is negligible ($\rho = 0.08$). A more relevant-sounding
answer is thus no more likely to be a correct one.
Because a fluent, well-structured, superficially responsive answer is exactly
what a non-expert user is least equipped to second-guess, this failure mode is
particularly problematic in a legal setting. It also carries a concrete
methodological warning. Relevancy- or coherence-based signals---often proposed as
cheap, reference-free proxies for answer quality---are poor proxies for legal
correctness and should not be used to gate LLM outputs in high-stakes legal
workflows.

\subsection{Closed-question accuracy overstates reliability but ranks models well}
Question format shapes the \emph{level} of measured reliability while largely
preserving the \emph{ranking} of models. Closed-question accuracy and free-text
correctness are strongly rank-correlated ($\rho = 0.94$), so inexpensive
multiple-choice evaluation is a useful screening tool for model selection.
However, the two differ considerably in absolute terms---a model scoring \num{0.86}
on multiple choice may be correct on fewer than half of its free-text
answers---so closed-question accuracy should be read as an \emph{upper bound} on
practical reliability, not an estimate of it.

\subsection{Where and why models fail}
Two structural patterns indicate where reliability breaks down. By area
(Figure~\ref{fig:heatmap}), the models perform worst on procedural and
administrative law (mean accuracy \num{0.71}) and best on labor, tax, and civil
law (\num{0.79}--\num{0.81}); the ordering of models is nonetheless stable across
areas, so weaker models degrade everywhere rather than trading strengths. By
complexity (Figure~\ref{fig:complexity-res}, for semi-open items), correctness
follows an inverted-U: it peaks at \emph{medium} complexity and is \emph{lowest}
at \emph{high} complexity---where conflict-of-norms and
systematic-interpretation tasks tax legal reasoning---with a secondary dip at
\emph{low} complexity, where items demand verbatim normative existence, exact
definitions, and precise citations. That models stumble on the seemingly easiest, citation-heavy
items is telling, since it points to parametric memorization of Colombian legal
text that is approximate rather than exact, the same brittleness that surfaces as
citation hallucination in the literature \citep{dahl2024}. The sub-task breakdown
(Figure~\ref{fig:subtask}) sharpens this: the very worst low-complexity sub-tasks
are identifying the deciding judge and confirming a norm's existence---tasks that
require an \emph{exact} fact rather than reasoning---whereas free-text quality is
weakest for administrative law in particular (Figure~\ref{fig:freetext-heatmap}).
Both patterns suggest that grounding answers in authoritative Colombian
sources---rather than relying on parametric recall---is a promising route to
higher reliability.

\subsection{Implications}
For legal practice and education in Colombia, our results support a
human-in-the-loop posture in which current LLMs can accelerate first drafts and
issue spotting but require expert verification, especially for the
citation-precise and reasoning-intensive tasks where they are weakest. Methodologically, the study
shows the value of an expert-validated, jurisdiction-specific benchmark that
separates recognition (closed questions) from generation (free text) and reports
format-appropriate metrics over explicit subsets---a design that reveals the
fluency--correctness gap that a single aggregate score would hide.

\subsection{Limitations}
\label{sec:limitations}
Several limitations qualify these findings. \emph{Metrics.} Free-text correctness
is measured with reference-anchored, two-judge-averaged RAGAS
(Section~\ref{sec:metrics})---an LLM-as-a-judge setup that, being anchored to a
human-written reference and averaged over two models, aligns well with human
judgment \citep{zheng2023judging,salinas2026lawprofs}, though like any automatic metric it is not a
full substitute for expert review. Full human grading of all
roughly \num{11055} free-text answers was not feasible; we instead validated the
scoring on a stratified subset, where the free-text ranking agrees across RAGAS,
an independent rubric-based judge, and human experts
(Section~\ref{sec:res-validation}). That validation also bounds what the
automatic scores support: they are reliable for \emph{ordering} models but read
about half a point low in \emph{absolute} terms, and the rubric judge's own
limitations---a single LLM judge of one family (no inter-judge reliability
estimate), no repeated items (no intra-rater reliability), no documented
calibration round, and roughly \num{21} validated cases per model---mean the bias
correction is coarse, and a second judge of a different family and a larger
human-validation sample would refine it. Two further metric caveats: BERTScore
proved nearly constant across models (\num{0.699}--\num{0.717}) and contributed
little discriminative signal---its base-model configuration should be reported in
full for reproducibility---and reference-based scoring can under-credit
legally valid answers phrased differently from the reference. \emph{Data.} The
open-ended subset is small (\num{55} items), two of the
twelve planned legal areas (international and environmental law) are absent, and
the benchmark reflects Colombian law at a fixed point in time, whereas both the
law and the models evolve. \emph{Data-quality reporting.} We report the
attrition and editing volume of the qualification protocol
(Section~\ref{sec:qa}), but the protocol did not include independent double
coding of items with a formal inter-annotator agreement statistic; double-coding
a stratified subset to estimate that agreement is a worthwhile addition for
future iterations. \emph{Evaluation.} Each model was queried once, at
temperature~0 with a single prompt version per question type, so we do not
characterize sensitivity to decoding randomness or prompt phrasing. Addressing
these limitations---a second judge of a different family, a larger open-ended
set, full area coverage, and prompt/decoding robustness analyses---is the natural
next step for this benchmark.

\section{Conclusion}
\label{sec:conclusion}
We presented the first large-scale, expert-validated evaluation of LLM
reliability on the Colombian legal system, comprising a benchmark of \num{1042}
items across ten areas of law and three question formats, built through a
human-in-the-loop pipeline with multi-stage expert review, together with a
reproducible evaluation of
\num{15} contemporary models. Our main finding is a gap between apparent and
actual competence. Frontier models answer structured, closed questions well, but on
free-text legal analysis their factual correctness remains low
(at most \num{0.45}) and also decoupled from how relevant and fluent
their answers appear. Closed-question accuracy tracks model ranking faithfully
yet overstates practical reliability, and performance degrades predictably in the
legal areas where the models perform worst and at the extremes of question
complexity.

These results argue against unsupervised use of current LLMs to answer questions
about the Colombian legal system and in favor of a human-in-the-loop posture in
which experts verify
model output---especially citation-precise and reasoning-intensive tasks. They
also point to source grounding, rather than parametric recall, as the most
promising direction for improvement, given that the hardest failures stem from
approximate parametric memory of specific legal texts. Future work includes human validation
of free-text scoring, expanding the open-ended subset and completing coverage of
all twelve legal areas, evaluating retrieval-augmented and Spanish- or
Colombian-adapted models, and analyzing robustness to prompt and decoding
variation. We release the benchmark construction and evaluation pipeline to
enable reproducible, jurisdiction-specific assessment of legal LLMs.

\appendix
\begin{center}
  \Large\bfseries Appendices
\end{center}
\section{Example Questions}
\label{app:examples}
The following items illustrate the three question types and, for the semi-open
type, the three complexity levels. Original items are in
Spanish; English glosses are provided for the reader. Answers are abbreviated.

\paragraph{Closed (multiple-choice) --- Commercial law.}
\emph{Under the Colombian Commercial Code, the consequence of stipulating against
mandatory rules (\emph{normas imperativas}) is:}
\begin{enumerate}[label=(\alph*),leftmargin=2.4em]
  \item Relative nullity, since the parties may ratify by virtue of the principle
  of freedom of contract.
  \item Non-existence, without need of a judicial declaration.
  \item \textbf{Absolute nullity.} \quad(\emph{correct})
  \item Non-existence, but decreed by the competent judge under the Commercial Code.
\end{enumerate}
\emph{Legal basis:} Article 889 of the Colombian Commercial Code.

\paragraph{Semi-open (low complexity, normative existence) --- Administrative law.}
\emph{Question:} Does the Colombian legal order contain a rule regulating the
right to petition (\emph{derecho de petici\'on})? \\
\emph{Reference answer:} Yes---Article 23 of the Political Constitution and Law
1755 of 2015, which regulates the fundamental right to petition.

\paragraph{Semi-open (medium complexity, temporal validity) --- Treaty / constitutional law.}
\emph{Question:} When did the Free Trade Agreement between Colombia and the
United States enter into force? \\
\emph{Reference answer:} On \num{15} May \num{2012}, once the legislative and
constitutional approval requirements had been met in both states. In Colombia the
agreement was incorporated by Law 1143 of 2007---held constitutional by the
Constitutional Court in ruling C-780 of 2008, and its amending protocol (Law 1166
of 2007) in ruling C-751 of 2008---and Decree 993 of 15 May 2012 gave effect to
it.

\paragraph{Semi-open (high complexity, normative conflict) --- Commercial law.}
\emph{Question:} Is there a normative conflict between Article 1324(1) of the
Commercial Code---which grants the commercial agent a payment upon termination of
the contract---and Articles 15--16 of the Civil Code---which allow the waiver of
rights of individual interest unless expressly prohibited? How is this apparent
tension between the special commercial rule and the general principles of private
autonomy resolved? \\
\emph{Reference answer (abridged):} The Supreme Court of Justice (judgment of
19 October 2011, file 11001-3103-032-2001-00847-01) held that there is no real
contradiction but complementarity. Applying the principle of speciality, Article
1324 does not expressly prohibit waiver; read systematically with Article 4 of
the Commercial Code and Articles 15--16 of the Civil Code, and in harmony with
the constitutional principles of contractual freedom, private autonomy, and
business initiative (Articles 16, 58, and 333 of the Constitution), the agent's
payment is a \emph{natural}---not essential or public-order---element of the
contract, so the parties may modify or waive it, provided they observe good faith
and the requirements of contractual validity.

\paragraph{Open-ended (IRAC) --- Criminal / data-protection law.}
\emph{Question:} A casuistic prompt describing a ransomware cyberattack against
a health-insurance provider (EPS) and a subsequent international transfer of
personal data; the model must analyze the criminal liability that may arise.
\emph{Reference answer (abridged):} A structured IRAC analysis identifying a
heterogeneous concurrence of offenses under Articles 269A, 269B, 269F and 244 of
the Criminal Code (added by Law 1273 of 2009), the protected legal interests and
the active/passive subjects, and the aggravating circumstances of Article 269H,
grounded in Law 1581 of 2012 on data protection.

\medskip
\noindent\textbf{Illustrative hardest items.} The following three semi-open items
are among the ten lowest-scoring in the benchmark
(Table~\ref{tab:hardest-freetext}); they show that the sharpest failures are
tasks of \emph{exact recall}---an authority's name or a precise article---rather
than legal reasoning.

\paragraph{Exact attribution --- Constitutional law (ruling on the judicial use of AI).}
\emph{Question:} What is the name of the reporting magistrate
(\emph{Magistrado Ponente}) of Constitutional Court ruling T-323 of 2024? \\
\emph{Reference answer:} Juan Carlos Cort\'es Gonz\'alez. \\
\emph{Mean correctness \num{0.07}.} The ruling---a landmark on the adequate use of
AI in judicial decisions---is recent, and the task is pure attribution recall; no
model reliably recovers the magistrate's name.

\paragraph{Exact citation --- Procedural law (conciliation).}
\emph{Question:} Which article of the law sets out the requirements to be a
conciliator? \\
\emph{Reference answer:} Article 28 of Law 2220 of 2022 (the General Conciliation
Statute). \\
\emph{Mean correctness \num{0.09}.} The item requires the exact article of a
recent statute; models approximate the citation rather than reproduce it.

\paragraph{Recent case law --- Tax law.}
\emph{Question:} What was the holding (\emph{sentido del fallo}) of Council of
State ruling 26644 of 2024? \\
\emph{Reference answer:} It recognized that taxes paid abroad on income of
national source may be deducted as an expense from income tax, since---being
related to national-source income---they are not creditable against that same tax. \\
\emph{Mean correctness \num{0.10}.} The decision post-dates the models' training
horizon, so the specific holding lies beyond their parametric knowledge.

\section{Prompt Templates}
\label{app:prompts}
Each question type was answered under a dedicated, version-controlled prompt
template. The models were prompted in Spanish; the verbatim templates are
reproduced below (the required JSON response keys are shown in typewriter font)
and are released with the benchmark pipeline. Every template casts the model as
an expert in Colombian law and requires it to ground its answer in the applicable
Colombian legal source and to return a structured JSON response so that answers
can be parsed and scored automatically.

\paragraph{Closed (single-best-answer multiple-choice).}
\begin{quote}\small
Eres un experto en derecho colombiano. Tu tarea es responder una pregunta de
selecci\'on m\'ultiple sobre derecho colombiano.

Se te proporcionar\'a una pregunta con exactamente 4 opciones de respuesta
etiquetadas A, B, C y D. Debes: (1)~analizar cada opci\'on a la luz del
ordenamiento jur\'idico colombiano vigente; (2)~seleccionar la opci\'on correcta;
(3)~explicar de forma clara y concisa por qu\'e la opci\'on elegida es correcta,
citando la norma, art\'iculo, ley o jurisprudencia colombiana aplicable;
(4)~indicar brevemente por qu\'e las dem\'as opciones son incorrectas.

Responde en formato JSON con las llaves: \texttt{respuesta\_correcta} (letra de
la opci\'on), \texttt{justificacion} y \texttt{descarte\_opciones} (una breve
raz\'on de descarte por cada opci\'on incorrecta).
\end{quote}
\noindent\emph{English translation.}
\begin{quote}\small
You are an expert in Colombian law. Your task is to answer a multiple-choice
question about Colombian law.

You will be given a question with exactly 4 answer options labelled A, B, C, and
D. You must: (1)~analyze each option in the light of the Colombian legal order in
force; (2)~select the correct option; (3)~explain clearly and concisely why the
chosen option is correct, citing the applicable Colombian norm, article, statute,
or case law; (4)~briefly indicate why the remaining options are incorrect.

Respond in JSON format with the keys: \texttt{respuesta\_correcta} (the option
letter), \texttt{justificacion}, and \texttt{descarte\_opciones} (a brief reason
for discarding each incorrect option).
\end{quote}

\paragraph{Semi-open.}
\begin{quote}\small
Eres un experto en derecho colombiano. Tu tarea es responder una pregunta
semiabierta sobre derecho colombiano.

Se te proporcionar\'a una pregunta que requiere una respuesta guiada y
estructurada. Debes: (1)~identificar los conceptos jur\'idicos clave involucrados
en la pregunta; (2)~proporcionar una respuesta concisa de entre 3 y 5 oraciones,
fundamentada en el ordenamiento jur\'idico colombiano; (3)~incluir las palabras o
frases clave que son esenciales para una respuesta correcta; (4)~citar la norma,
art\'iculo, ley o jurisprudencia colombiana relevante que sustenta tu respuesta.

Responde en formato JSON con las llaves: \texttt{respuesta},
\texttt{palabras\_clave} (lista de t\'erminos esenciales) y
\texttt{referencia\_legal} (norma o art\'iculo citado).
\end{quote}
\noindent\emph{English translation.}
\begin{quote}\small
You are an expert in Colombian law. Your task is to answer a semi-open question
about Colombian law.

You will be given a question that calls for a guided, structured answer. You
must: (1)~identify the key legal concepts involved in the question;
(2)~provide a concise answer of between 3 and 5 sentences, grounded in the
Colombian legal order; (3)~include the words or phrases that are essential to a
correct answer; (4)~cite the relevant Colombian norm, article, statute, or case
law supporting your answer.

Respond in JSON format with the keys: \texttt{respuesta},
\texttt{palabras\_clave} (a list of essential terms), and
\texttt{referencia\_legal} (the cited norm or article).
\end{quote}

\paragraph{Open-ended.}
\begin{quote}\small
Eres un experto en derecho colombiano. Tu tarea es responder una pregunta abierta
de an\'alisis jur\'idico sobre derecho colombiano.

Se te proporcionar\'a una pregunta que requiere un an\'alisis jur\'idico profundo
y argumentado. Debes: (1)~identificar el marco normativo aplicable, incluyendo
leyes, decretos, art\'iculos constitucionales o c\'odigos relevantes del
ordenamiento jur\'idico colombiano; (2)~desarrollar un an\'alisis jur\'idico
detallado de entre 5 y 8 oraciones que aborde los argumentos principales y las
distintas posturas doctrinales si las hay; (3)~referenciar jurisprudencia
relevante de la Corte Constitucional, la Corte Suprema de Justicia o el Consejo
de Estado de Colombia, si aplica; (4)~formular una conclusi\'on clara que resuma
la postura jur\'idica fundamentada.

Responde en formato JSON con las llaves: \texttt{marco\_normativo},
\texttt{analisis}, \texttt{jurisprudencia} y \texttt{conclusion}.
\end{quote}
\noindent\emph{English translation.}
\begin{quote}\small
You are an expert in Colombian law. Your task is to answer an open-ended legal-analysis
question about Colombian law.

You will be given a question that requires a deep, reasoned legal analysis. You
must: (1)~identify the applicable normative framework, including relevant
statutes, decrees, constitutional articles, or codes of the Colombian legal
order; (2)~develop a detailed legal analysis of between 5 and 8 sentences
addressing the main arguments and the different doctrinal positions, if any;
(3)~reference relevant case law of the Constitutional Court, the Supreme Court of
Justice, or the Council of State of Colombia, where applicable; (4)~state a clear
conclusion that summarizes the reasoned legal position.

Respond in JSON format with the keys: \texttt{marco\_normativo} (normative
framework), \texttt{analisis} (analysis), \texttt{jurisprudencia} (case law), and
\texttt{conclusion}.
\end{quote}

Note that the open-ended template requests a four-part structure
(\texttt{marco\_normativo}, \texttt{analisis}, \texttt{jurisprudencia},
\texttt{conclusion}) that maps onto the Rule--Application--Conclusion of the IRAC
reference answers without naming the IRAC method explicitly, as discussed in
Section~\ref{sec:prompting}.

\bibliographystyle{plainnat}
\bibliography{references}

\end{document}